\pdfoutput=1
\documentclass[11pt]{article}

\usepackage[final]{acl}

\usepackage{times}
\usepackage{latexsym}

\usepackage[T1]{fontenc}
\usepackage[utf8]{inputenc}

\usepackage{microtype}

\usepackage{inconsolata}

\usepackage{graphicx}

\usepackage{amsmath}
\usepackage{amsthm}
\usepackage{bm}
\usepackage{amsfonts}
\usepackage{multirow}
\usepackage{diagbox}
\usepackage{caption}
\usepackage{subcaption}
\usepackage{booktabs}
\usepackage{tabularx}
\usepackage{graphicx}
\usepackage{cleveref}
\usepackage{verbatim}
\usepackage{pifont}

\usepackage{paralist}

\title{Inference-Time Selective Debiasing to Enhance Fairness in \\ Text Classification Models}

\author{\bf Gleb Kuzmin\textsuperscript{2,4} \qquad
  Neemesh Yadav\textsuperscript{5} \qquad
  Ivan Smirnov\textsuperscript{3,4} \qquad \\
  \bf Timothy Baldwin\textsuperscript{1,6} \qquad
  Artem Shelmanov\textsuperscript{1} \\
\textsuperscript{1}MBZUAI ~~
\textsuperscript{2}AIRI ~~
\textsuperscript{3}RUDN University ~~
\textsuperscript{4}FRC CSC RAS \\
\textsuperscript{5}IIIT Delhi ~~
\textsuperscript{6}The University of Melbourne
\\
\href{mailto:kuzmin@airi.net}{kuzmin@airi.net} ~~ \href{mailto:neemesh20529@iiitd.ac.in}{neemesh20529@iiitd.ac.in} ~~ \href{mailto:ivs@isa.ru}{ivs@isa.ru}\\
\href{mailto:artem.shelmanov@mbzuai.ac.ae}{\{timothy.baldwin, artem.shelmanov\}@mbzuai.ac.ae}
}

\begin{document}

\maketitle

\begin{abstract}
We propose selective debiasing -- an inference-time safety mechanism designed to enhance the overall model quality in terms of prediction performance and fairness, especially in scenarios where retraining the model is impractical. The method draws inspiration from selective classification, where at inference time, predictions with low quality, as indicated by their uncertainty scores, are discarded.
In our approach, we identify the potentially biased model predictions and, instead of discarding them, we remove bias from these predictions using LEACE -- a post-processing debiasing method. 
To select problematic predictions, we propose a bias quantification approach based on KL divergence, which achieves better results than standard uncertainty quantification methods. Experiments on text classification datasets with encoder-based classification models demonstrate that selective debiasing helps to reduce the performance gap between post-processing methods and debiasing techniques from the at-training and pre-processing categories.\footnote{The code is available online at \url{https://github.com/glkuzi/selective-debiasing}}
\end{abstract}

\section{Introduction}

Fairness is an important safety characteristic of a machine learning (ML) model, representing the model's ability to classify instances without discrimination based on various sensitive attributes, such as race, gender, and age \cite{blodgett-etal-2020-language}.
For the past few years, numerous works have investigated and promoted fairness, and a variety of fairness definitions have been proposed \cite{blodgett-etal-2020-language,han-etal-2022-fairlib}. One prominent type of fairness is group fairness, also known as the equal opportunity criterion, which reflects the inequality of opportunities across different groups \cite{han-etal-2022-balancing}. The inequality in the model predictions usually comes from inadequate or biased training data, and to address this problem and achieve better fairness, researchers have proposed various debiasing techniques \cite{li-etal-2018-towards,han-etal-2021-diverse,han-etal-2022-balancing,NEURIPS2023_d066d21c,kuzmin-etal-2023-uncertainty}. The majority of these techniques assume that one has access to the complete training data and the ability to retrain the model from scratch using some special loss function or reweighting the training instances. However, there are many situations when this assumption does not hold. There is a need for inference-time safety mechanisms that protect users from inadequate model behavior.

Inference-time safety mechanisms are primarily associated with uncertainty quantification (UQ) techniques 
\cite{gal2016dropout} 
and selective classification \cite{geifman2017selective,xin-etal-2021-art,vazhentsev-etal-2022-uncertainty,vazhentsev-etal-2023-hybrid}. Selective classification aims to enhance the reliability of ML-based applications by abstaining from unreliable predictions with high uncertainty. We suggest that the same approach could be applied to increase fairness.

In this work, we propose an inference-time safety mechanism that aims to increase the overall quality of models in terms of prediction performance and fairness in situations when model retraining is prohibitive. 
We call this approach \textit{selective debiasing}. Instead of rejecting predictions of selected instances as in selective classification, we apply to them inference-time debiasing using post-processing debiasing techniques. 
To the best of our knowledge, this style of approach is novel to the NLP community.

Our main contributions are as follows: 
\begin{compactitem}
    \item We propose selective debiasing, an inference-time safety mechanism that aims to improve both the performance and fairness of model predictions by applying a post-processing debiasing method to only a selected subset of predictions.

    \item We suggest a scoring criterion that aims to select the most unreliable and biased predictions. Experiments demonstrate that this scoring criterion is generally better than UQ techniques in selective debiasing.
\end{compactitem}

\section{Background}
\label{sec:background}

Debiasing techniques can be categorized into three groups: at-training, pre-processing, and post-processing \cite{han-etal-2022-fairlib}. 

\paragraph{At-training and pre-processing methods.} One of the most popular at-training methods is adversarial training (Adv) \cite{li-etal-2018-towards}. It aims to solve a minimax game between minimizing the loss for the primary task and maximizing the loss for predicting the protected attribute. The diverse adversaries method (DAdv) \cite{han-etal-2021-diverse} extends Adv by using an ensemble of multiple diverse discriminators instead of just one. In the pre-processing category, one of the most remarkable methods is Balanced Training with Equal Opportunity (BTEO) \cite{han-etal-2022-balancing}. It rebalances the dataset to minimize the True Positive Rate (TPR) gap between two protected groups. In the same category, Balanced Training with Joint balance (BTJ) \cite{neurips-lahoti-btj} aims to improve the worst-case performance over all unobserved protected groups by focusing on the computationally identifiable regions of error.

\paragraph{Post-processing methods.} There are two well-known approaches to post-processing debiasing: Iterative Null-space Projection (INLP) \cite{ravfogel-etal-2020-null} and LEAst-squares Concept Erasure (LEACE) \cite{NEURIPS2023_d066d21c}.

INLP is an iterative method that involves finding an orthogonal projection of a linear classifier matrix, which is initially learned to predict protected attributes from representations (e.g.\ hidden states of the standard model). This orthogonal projection is then iteratively used to remove all relevant information from these representations, which was used by the classifier to predict protected attributes.

LEACE is a 
concept erasure technique that renders representations impervious to the prediction of a specific concept while minimizing changes to the original representations. To construct a transformation matrix, it first whitens the data by equalizing the variance across all directions in the representation space. Next, the data is orthogonally projected onto the subspace that captures correlations between representations and protected attributes. Finally, the data is unwhitened using the same covariance matrix. This resulting transformation matrix is subtracted from the original representations (see the formal definition for LEACE in \Cref{sec:deb_tech_advanced}). 

\vspace{0.3cm}

At-training and pre-processing methods require retraining the model from scratch and access to the whole training set. They also cannot be selectively applied to a subset of predictions. Post-processing techniques do not involve changes to the model itself, can be trained on a subset of data, and can be applied to predictions selectively. However, their performance is usually worse.

In our work, we propose a method that combines the advantages of both post-processing and at-training / pre-processing methods. While it does not need access to the whole training dataset or retraining the model from scratch, it also has better performance than the standard post-processing techniques.

\section{Proposed Method}
We propose a selective approach, based on applying debiasing only to predictions with the highest bias score. This section introduces the general concept of selective debiasing and presents the bias quantification method underlying this approach.

\paragraph{Selective debiasing.}

Selective classification is a widely recognized safety mechanism that safeguards against using unreliable model predictions. In this approach, predictions flagged as unreliable due to high uncertainty scores are handled differently, e.g. they are rejected or are escalated to human operators for further review.

Instead of rejecting instances completely as in selective classification, we apply debiasing to selected predictions.  
In particular, we identify the potentially most biased instances using a bias quantification method $\mathcal{B}(x_i, p_i)$ and replace the original prediction $p_i=f(x_i)$ with a prediction debiased using a post-processing method $d$: $\hat{p_i}=d(f(x_i))$:
\begin{equation}
    \bar{p_i}=
    \begin{cases}
        p_i = f(x_i), \text{ if } \mathcal{B}(x_i, p_i) < h \\
        \hat{p_i}= d(f(x_i)), \text{ if } \mathcal{B}(x_i, p_i) \geq h, 
    \end{cases}
\end{equation}
where $h$ is a predefined threshold selected on a validation set.

We note that the proposed approach is different from the standard post-processing debiasing methods since we change predictions for only some instances. While debiasing all predictions might significantly reduce model performance, modifying only predictions likely to be of low quality or biased is less risky in terms of worsening outcomes and has the potential to correct errors. Such an approach also allows tuning the accuracy--fairness trade-off for debiasing methods \cite{han-etal-2022-fairlib,kuzmin-etal-2023-uncertainty}.

\begin{figure}[t!]
    \centering
    \includegraphics[width=0.45\textwidth,trim={0.4cm 0.4cm 0.35cm 0.35cm},clip]{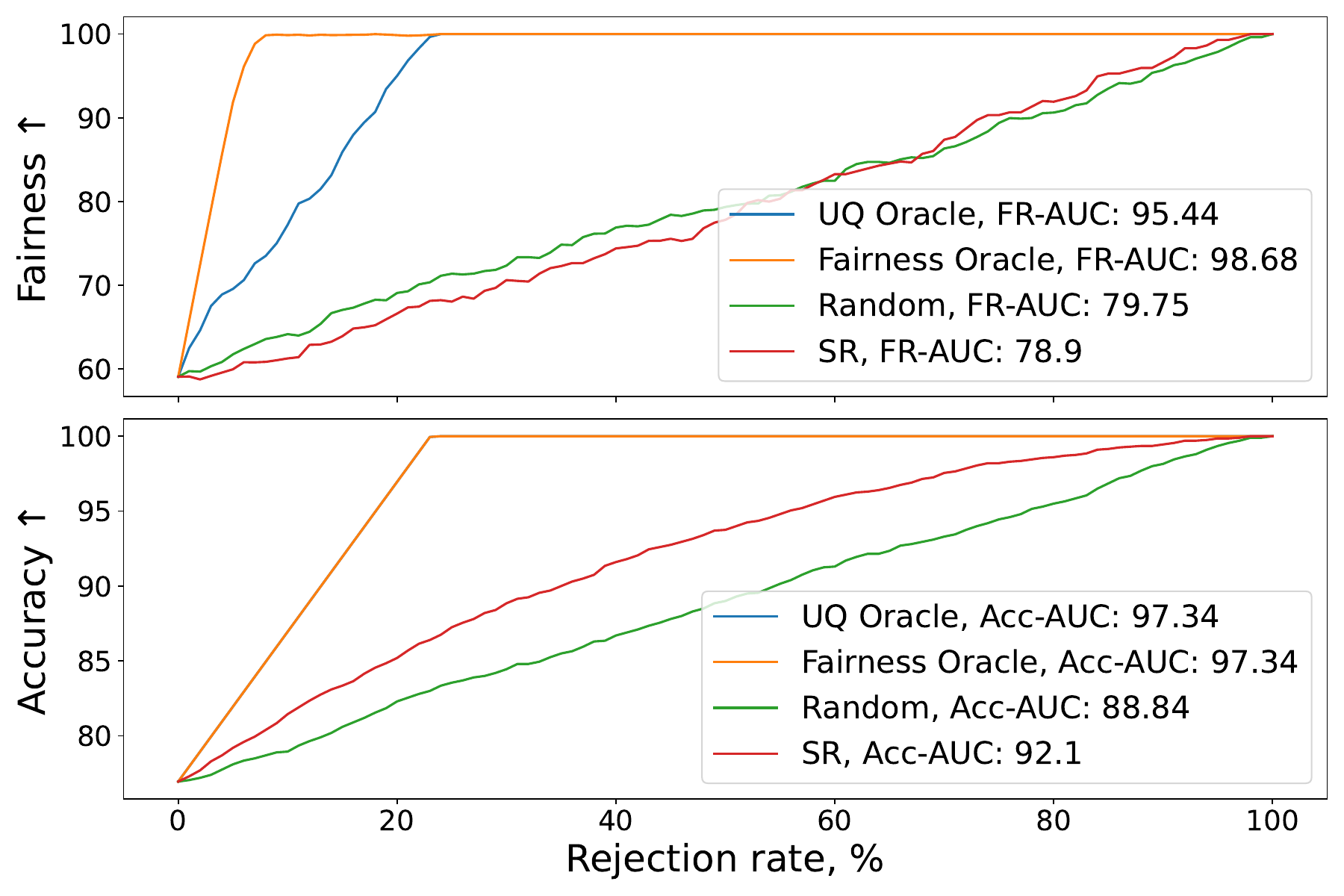}
    \caption{Rejection results for fairness and accuracy with oracle scores on a synthetic dataset with a LogReg model; the FR-AUC and Acc-AUC are the areas under fairness-- and accuracy--rejection curves correspondingly. The details are presented in \Cref{sec:fairness_oracle}.}
    \label{fig:synth_oracle}
\end{figure}

\paragraph{Bias quantification method.}
Selective classification is usually based on UQ methods. However,
uncertainty on its own does not reflect the presence of bias; it simply
highlights potentially erroneous predictions. \Cref{fig:synth_oracle} presents a motivational example. It shows the rejection plots for oracle rejection strategies in selective classification for both accuracy and fairness (see the exact definition of fairness in \Cref{sec:fairness_metric}). We can see that the fairness oracle outperforms the UQ oracle in terms of fairness while keeping the same performance in terms of accuracy. These results illustrate that it is possible to improve fairness without penalty to accuracy by changing the order of instances being eliminated, i.e. using a different selection criterion.

Consider a multi-label classification model with classes $c \in C$. To quantify how biased a model prediction is for a given instance, we suggest using the Kullback-Leibler (KL) divergence  \cite{kullback1951information} between the originally predicted probability distribution $p_{i}^{c}$ and distribution $\hat{p_{i}}^{c}$ after debiasing:
\begin{equation}\label{eq:kl}
    %\small
    \textstyle
    \mathcal{B}_{KL}^{i}=\sum_{c\in C}p_{i}^{c}\log\left(\frac{p_{i}^{c}}{\hat{p_{i}}^{c}}\right).
\end{equation}

KL divergence measures the difference in predictions between the standard and the debiased model. The greater the difference, the more information about the protected attribute is removed from the original representation of the instance. This approach could be used with various post-processing methods. In particular, we suggest using LEACE, but also present results with INLP.

Note that applying a post-processing method to a model is a matter of one or two matrix multiplications. An additional prediction step requires inferring only the last layer of a model, which is very fast. Therefore, the runtime overhead introduced by bias quantification is very small (see \Cref{sec:comp_stats}).

\section{Experiments}
\subsection{Experimental Setup}

\paragraph{Datasets.}
For our experiments, we use two English text classification datasets that, in addition to target variables, provide explicit protected attributes.  
The first is MOJI \cite{blodgett-etal-2016-demographic}, a dataset for sentiment analysis with a binary class (``happy'' and ``sad'') and a binary protected attribute, which corresponds to the author's ethnicity (African American English (AAE) vs.\ Standard American English (SAE)). The second is a version of the widely used BIOS dataset \cite{de2019bias} for occupation classification with a binary gender as the protected attribute. BIOS-2 \cite{subramanian-etal-2021-evaluating} is a two-class subsample of the original BIOS dataset with a highly-skewed joint distribution of classes and protected attribute values. As it has been shown to be beneficial to report results for both ``balanced'' and ``imbalanced'' versions of datasets \cite{kuzmin-etal-2023-uncertainty}, we conduct experiments on both versions. Detailed information and statistics of the datasets are presented in \Cref{sec:data_stats}. Due to the limited availability of datasets with annotated protected attributes, most research on debiasing and fairness has been conducted on these few datasets \cite{han-etal-2022-fairlib}.

\begin{table*}[!ht] 
\centering
\resizebox{\textwidth}{!}{\begin{tabular}{ll||c|cc|cc||ccc||ccc}
\toprule
\multicolumn{2}{c||}{\textbf{Debiasing method type}} & \textbf{No debiasing} & \multicolumn{2}{c|}{\textbf{At-training}} & \multicolumn{2}{c||}{\textbf{Pre-processing}} & \multicolumn{6}{c}{\textbf{Post-processing \& Selective}} \\
\toprule
\textbf{Dataset} & \textbf{Metric} & \textbf{Standard} & \textbf{Adv} & \textbf{DAdv} & \textbf{BTEO} & \textbf{BTJ} & \begin{tabular}[c]{@{}l@{}}\textbf{LEACE-}\\ \textbf{last}\\ \end{tabular} & \begin{tabular}[c]{@{}l@{}}\textbf{LEACE-}\\ \textbf{last+SR,}\\ \textbf{opt.} \textbf{perc.}\end{tabular} & \begin{tabular}[c]{@{}l@{}}\textbf{LEACE-}\\ \textbf{last+KL,}\\ \textbf{opt.} \textbf{perc.}\end{tabular} & \begin{tabular}[c]{@{}l@{}}\textbf{LEACE-}\\ \textbf{cls}\\ \end{tabular} & \begin{tabular}[c]{@{}l@{}}\textbf{LEACE-}\\ \textbf{cls+SR,}\\ \textbf{opt.} \textbf{perc.}\end{tabular} & \begin{tabular}[c]{@{}l@{}}\textbf{LEACE-}\\ \textbf{cls+KL,}\\ \textbf{opt.} \textbf{perc.}\end{tabular}\\
\midrule
\multirow[c]{4}{*}{\begin{tabular}[c]{@{}l@{}}MOJI\\ imbalanced\\ \end{tabular}} & Fairness $\uparrow$ & 61.8\tiny{$\pm$0.7} & 73.7\tiny{$\pm$0.6} & 73.4\tiny{$\pm$0.4} & \textbf{75.2\tiny{$\pm$0.6}} & 74.8\tiny{$\pm$0.6} & \textbf{75.8\tiny{$\pm$2.6}} & 68.6\tiny{$\pm$1.4} & 75.7\tiny{$\pm$0.8} & 75.2\tiny{$\pm$3.0} & 68.4\tiny{$\pm$1.1} & \underline{\textbf{77.2\tiny{$\pm$0.7}}}\\
 & Accuracy $\uparrow$ & \underline{\textbf{79.1\tiny{$\pm$0.7}}} & 72.0\tiny{$\pm$0.7} & 72.4\tiny{$\pm$0.5} & 73.6\tiny{$\pm$0.6} & 73.2\tiny{$\pm$0.4} & 68.3\tiny{$\pm$2.6} & \textbf{77.6\tiny{$\pm$0.9}} & 72.7\tiny{$\pm$1.2} & 66.8\tiny{$\pm$3.0} & \textbf{77.6\tiny{$\pm$1.0}} & 71.8\tiny{$\pm$1.2} \\
 & DTO $\downarrow$ & 43.6\tiny{$\pm$0.6} & 38.4\tiny{$\pm$0.5} & 38.3\tiny{$\pm$0.4} & \underline{\textbf{36.2\tiny{$\pm$0.1}}} & 36.7\tiny{$\pm$0.4} & 39.9\tiny{$\pm$3.6} & 38.6\tiny{$\pm$0.7} & \textbf{36.6\tiny{$\pm$0.5}} & 41.4\tiny{$\pm$4.1} & 38.8\tiny{$\pm$0.6} & \underline{\textbf{36.2\tiny{$\pm$1.2}}}  \\
 & FF-score $\uparrow$ & 69.4\tiny{$\pm$0.4} & 72.8\tiny{$\pm$0.4} & 72.9\tiny{$\pm$0.3} & \underline{\textbf{74.4\tiny{$\pm$0.1}}} & \textbf{74.0\tiny{$\pm$0.3}} & 71.8\tiny{$\pm$2.6} & 72.8\tiny{$\pm$0.5} & \textbf{74.1\tiny{$\pm$0.3}} & 70.8\tiny{$\pm$3.0} & 72.7\tiny{$\pm$0.4} & \underline{\textbf{74.4\tiny{$\pm$0.8}}}  \\
\midrule
\multirow[c]{4}{*}{\begin{tabular}[c]{@{}l@{}}MOJI\\ balanced\\ \end{tabular}} & Fairness $\uparrow$ & 69.5\tiny{$\pm$0.2} & 83.8\tiny{$\pm$0.8} & 84.7\tiny{$\pm$1.5} & 85.5\tiny{$\pm$0.5} & \textbf{85.6\tiny{$\pm$0.6}} & 79.7\tiny{$\pm$3.9} & 77.1\tiny{$\pm$0.9} & \textbf{86.6\tiny{$\pm$0.5}} & 77.6\tiny{$\pm$4.2} & 77.0\tiny{$\pm$0.8} & \underline{\textbf{87.5\tiny{$\pm$0.5}}}  \\
 & Accuracy $\uparrow$ & 71.9\tiny{$\pm$0.4} & 74.0\tiny{$\pm$0.4} & 74.1\tiny{$\pm$0.6} & \underline{\textbf{74.8\tiny{$\pm$0.3}}} & 74.5\tiny{$\pm$0.4} & 73.6\tiny{$\pm$0.8} & \textbf{74.0\tiny{$\pm$0.3}} & 74.0\tiny{$\pm$0.2} & 73.0\tiny{$\pm$1.2} & \textbf{74.0\tiny{$\pm$0.4}} & 73.7\tiny{$\pm$0.5}  \\
 & DTO $\downarrow$ & 41.5\tiny{$\pm$0.4} & 30.7\tiny{$\pm$0.7} & 30.1\tiny{$\pm$0.7} & \underline{\textbf{29.0\tiny{$\pm$0.1}}} & 29.3\tiny{$\pm$0.4} & 33.4\tiny{$\pm$3.0} & 34.7\tiny{$\pm$0.7} & \textbf{29.3\tiny{$\pm$0.3}} & 35.2\tiny{$\pm$3.7} & 34.7\tiny{$\pm$0.6} & \textbf{29.1\tiny{$\pm$0.6}} \\
 & FF-score $\uparrow$ & 70.7\tiny{$\pm$0.3} & 78.6\tiny{$\pm$0.5} & 79.1\tiny{$\pm$0.6} & \textbf{79.8\tiny{$\pm$0.1}} & 79.6\tiny{$\pm$0.3} & 76.5\tiny{$\pm$2.2} & 75.5\tiny{$\pm$0.5} & \textbf{79.8\tiny{$\pm$0.2}} & 75.2\tiny{$\pm$2.6} & 75.5\tiny{$\pm$0.4} & \underline{\textbf{80.0\tiny{$\pm$0.4}}} \\
\midrule
\multirow[c]{4}{*}{\begin{tabular}[c]{@{}l@{}}BIOS-2\\ imbalanced\\ \end{tabular}} & Fairness $\uparrow$ & 90.4\tiny{$\pm$0.8} & \underline{\textbf{97.2\tiny{$\pm$0.8}}} & 96.4\tiny{$\pm$0.4} & 95.8\tiny{$\pm$1.0} & 96.6\tiny{$\pm$0.8} & 92.8\tiny{$\pm$9.3} & 93.0\tiny{$\pm$2.3} & \textcolor{gray}{\textbf{94.5\tiny{$\pm$4.4}}} & 77.3\tiny{$\pm$6.5} & 94.8\tiny{$\pm$2.3} & \textbf{96.7\tiny{$\pm$0.9}} \\
 & Accuracy $\uparrow$ & \underline{\textbf{96.7\tiny{$\pm$0.1}}} & 94.8\tiny{$\pm$0.4} & 95.0\tiny{$\pm$0.3} & 95.2\tiny{$\pm$0.3} & 95.0\tiny{$\pm$0.5} & 60.5\tiny{$\pm$3.6} & \textbf{94.6\tiny{$\pm$0.2}} & 92.0\tiny{$\pm$0.4} & 64.0\tiny{$\pm$5.5} & \textbf{94.6\tiny{$\pm$0.1}} & 93.2\tiny{$\pm$0.3} \\
 & DTO $\downarrow$ & 10.1\tiny{$\pm$0.7} & \underline{\textbf{5.9\tiny{$\pm$0.2}}} & 6.2\tiny{$\pm$0.2} & 6.5\tiny{$\pm$0.6} & 6.1\tiny{$\pm$0.3} & 41.3\tiny{$\pm$2.1} & \textbf{9.0\tiny{$\pm$1.7}} & \textcolor{gray}{10.3\tiny{$\pm$2.8}} & 43.4\tiny{$\pm$2.4} & 7.7\tiny{$\pm$1.7} & \textbf{7.6\tiny{$\pm$0.5}} \\
 & FF-score $\uparrow$ & 93.5\tiny{$\pm$0.4} & \underline{\textbf{96.0\tiny{$\pm$0.2}}} & 95.7\tiny{$\pm$0.1} & 95.5\tiny{$\pm$0.4} & 95.8\tiny{$\pm$0.2} & 72.8\tiny{$\pm$2.3} & \textbf{93.8\tiny{$\pm$1.2}} & \textcolor{gray}{93.2\tiny{$\pm$2.3}} & 69.6\tiny{$\pm$1.7} & 94.7\tiny{$\pm$1.2} & \textbf{94.9\tiny{$\pm$0.4}}  \\
\midrule
\multirow[c]{4}{*}{\begin{tabular}[c]{@{}l@{}}BIOS-2\\ balanced\\ \end{tabular}} & Fairness $\uparrow$ & 89.7\tiny{$\pm$0.6} & 97.8\tiny{$\pm$0.8} & \underline{\textbf{98.0\tiny{$\pm$0.8}}} & 95.9\tiny{$\pm$0.8} & 96.4\tiny{$\pm$0.3} & 90.6\tiny{$\pm$9.8} & 93.7\tiny{$\pm$2.6} & \textcolor{gray}{\textbf{94.6\tiny{$\pm$4.2}}} & 74.8\tiny{$\pm$9.2} & 96.6\tiny{$\pm$1.8} & \textbf{97.5\tiny{$\pm$0.9}} \\
 & Accuracy $\uparrow$ & 92.4\tiny{$\pm$0.3} & 91.9\tiny{$\pm$0.6} & 91.9\tiny{$\pm$1.5} & 92.6\tiny{$\pm$0.5} & \underline{\textbf{92.9\tiny{$\pm$0.6}}} & 49.9\tiny{$\pm$9.4} & \textbf{90.9\tiny{$\pm$1.3}} & 89.3\tiny{$\pm$1.8} & 63.8\tiny{$\pm$10.1} & \textbf{91.9\tiny{$\pm$0.7}} & 90.6\tiny{$\pm$1.3}  \\
 & DTO $\downarrow$ & 12.8\tiny{$\pm$0.6} & 8.5\tiny{$\pm$0.4} & 8.4\tiny{$\pm$1.4} & 8.5\tiny{$\pm$0.2} & \underline{\textbf{8.0\tiny{$\pm$0.6}}} & 52.4\tiny{$\pm$6.0} & \textbf{11.1\tiny{$\pm$2.4}} & \textcolor{gray}{12.4\tiny{$\pm$3.4}} & 46.0\tiny{$\pm$4.3} & \textbf{8.9\tiny{$\pm$1.4}} & 9.7\tiny{$\pm$1.5}  \\
 & FF-score $\uparrow$ & 91.1\tiny{$\pm$0.4} & 94.7\tiny{$\pm$0.1} & \underline{\textbf{94.9\tiny{$\pm$0.7}}} & 94.2\tiny{$\pm$0.2} & 94.6\tiny{$\pm$0.3} & 63.0\tiny{$\pm$4.6} & \textbf{92.3\tiny{$\pm$1.9}} & \textcolor{gray}{91.9\tiny{$\pm$2.9}} & 67.5\tiny{$\pm$3.0} & \textbf{94.2\tiny{$\pm$1.2}} & 93.9\tiny{$\pm$1.1} \\
 \bottomrule
\end{tabular}
}\caption{\label{tab:deb_performance} Comparison of debiasing methods and selective debiasing. The best results in the group are in bold, and the best results overall are underlined. The results are averaged over 5 random seeds. The gray color corresponds to the results with p-value > 0.05 with respect to standard model.}\end{table*}

\paragraph{Metrics.}
We employ several metrics to evaluate the predictive performance and fairness of the model. To evaluate the performance, we use accuracy. For fairness, we consider the widely used equal opportunity criterion \cite{NIPS2016_9d268236,han-etal-2022-balancing, han-etal-2022-fairlib}. We also use two aggregated metrics to evaluate the performance in terms of both accuracy and fairness. The first one is the distance to the optimal point (DTO) \cite{han-etal-2021-diverse}:
\begin{equation}\label{eq:dto}
    \scriptstyle
    \text{DTO}=\sqrt{(1-\text{Accuracy})^{2} + (1-\text{Fairness})^{2}}.
\end{equation}
The second one is the Fairness F-score (FF) -- a smoothed minimum of accuracy and fairness:
\begin{equation}\label{eq:fscore}
    \textstyle
    \text{FF-score}=\frac{2\cdot\text{Accuracy}\cdot\text{Fairness}}{\text{Accuracy}+\text{Fairness}}.
\end{equation}
Details of the equal opportunity fairness calculation are presented in Appendix \ref{sec:fairness_metric}.

\paragraph{Models.}
For the BIOS-2 dataset, we use BERT (``bert-base-cased'') \cite{devlin-etal-2019-bert}. For the MOJI dataset, we use the domain-specific BERTweet model \cite{nguyen-etal-2020-bertweet} which is good for processing data from social media sources. For both models, we add a three-layer MLP as a classification head, following \citet{han-etal-2022-fairlib}. Model hyperparameters are described in \Cref{sec:hyperpars}.

\paragraph{Baselines.}
We compare the proposed selective debiasing approach to inference-time debiasing of all predictions using LEACE and INLP, as well as to at-training and pre-processing debiasing techniques: Adv, DAdv, BTEO, BTJ.
We also compare the proposed KL-based bias quantification score with a UQ baseline: Softmax Response (SR: \citet{geifman2017selective}), calculated as $\mathcal{B}_{SR}(x_i)=1-\max_{c\in C} p^{c}_i$. 

\paragraph{Details of debiasing methods.}
Pre-processing and at-training debiasing methods were applied while training the model from scratch on the full dataset, whereas post-processing methods were trained using only 20\% of the data. The optimal threshold for selective debiasing was chosen based on the first 15\% of the validation set. ``LEACE-last'' in our experiments represents LEACE applied to the outputs of the last hidden layer of the classifier, while ``LEACE-cls'' is LEACE applied to each linear layer of the classification head of the model. The hyperparameters of debiasing methods are provided in \Cref{sec:hyperpars}.

\subsection{Results}
\Cref{tab:deb_performance} presents results for various at-training and pre-processing debiasing methods, post-processing debiasing methods, selective debiasing based on LEACE with SR, and selective debiasing using the proposed KL-based bias quantification score. Here, we show results only for the threshold that gives an optimal selection percentage. The full results with various selection percentages are presented in \Cref{sec:ablation_perc}. The results for selective debiasing using INLP are provided in \Cref{sec:ablation_perc}.

In the majority of cases, the best results are unsurprisingly achieved by at-training and pre-processing debiasing techniques, as these methods retrain the models from scratch on the full training data. Nevertheless, the proposed selective debiasing approach based on LEACE substantially enhances the results of inference-time debiasing using post-processing techniques in terms of metrics that take into account both fairness and performance: FF-score and DTO. Inference time debiasing becomes competitive with at-training and pre-processing techniques. For LEACE-cls with KL selection, selective debiasing even outperforms these methods on MOJI-balanced. The results in \Cref{tab:ablation_dist_leace_last,tab:ablation_dist_leace_cls,tab:ablation_dist_inlp} also show that selective debiasing consistently outperforms standard inference-time debiasing in terms of FF-score. 

LEACE-cls generally achieves better fairness than LEACE-last and slightly better joint fairness--performance in terms of DTO and FF-score.

When comparing the results of the proposed bias quantification method based on the KL distance with SR, we can see that our method notably outperforms SR on the MOJI datasets and is on par with SR on BIOS-2. We further explore other distance-based bias quantification methods (Euclidean and cosine distances) in Appendix \ref{sec:ablation_dist}. Results in \Cref{tab:ablation_dist_leace_last,tab:ablation_dist_leace_cls,tab:ablation_dist_inlp} show that in most cases, selection by KL works comparably or better than other distance-based measures. Moreover, KL scores are easier to compute than  distance-based scores.

\section{Conclusion and Future Work}

We proposed selective debiasing -- a new simple inference-time safety mechanism for increasing model performance and fairness. We showed that it is helpful in the case when re-training a model from scratch for better fairness is prohibitive or there is no access to full training data. Additionally, for the selection of problematic predictions, we suggest a bias quantification approach based on KL divergence that achieves better results than the standard UQ method. The proposed mechanism fills the gap for efficient techniques that can be applied at inference time and opens the door for safer ML-based systems. In future work, we aim to investigate a deeper integration between UQ and debiasing methods.

\section*{Limitations}
In this work, we considered only group fairness (equal opportunity criterion), where there exist many other fairness definitions. However, this research is focused particularly on group fairness, and the equal opportunity criterion is the metric of choice in previous work on the same datasets. During all experiments, we assume that we have access to the protected attributes, which is not always the case. But this is a common assumption for any work on debiasing; moreover, it is necessary for the calculation of the fairness metric. Finally, all of the experiments were conducted on the English language, but the used methods are language-independent, so we do not expect significant differences in results for other languages.

\section*{Ethical Considerations}
In this work, we consider group fairness and instance-level bias quantification. We used only publicly available datasets and models, and only for the intended use. In our research, we used protected attributes to apply debiasing methods and to compute metrics; however, this is necessary for all debiasing methods. To avoid possible harm, we used only attributes that users self-disclosed for the experiments.

\section*{Acknowledgments}
We appreciate the anonymous reviewers for their valuable suggestions that helped enhance this paper. This research was supported in part through the computational resources of HPC facilities at HSE University.

\bibliography{custom}

\clearpage
\appendix

\section{LEAst-Squares Concept Erasure}
\label{sec:deb_tech_advanced}
LEACE removes information about a concept $Z$ from the representation space $X$. To formally describe LEACE, we firstly introduce the following notions. Let $\mathbf{x}\in X$ be an instance from $X$ (e.g. embedding from the last layer in the case of LEACE-last), $\mathbf{\Sigma_{XX}}$ is the covariance matrix for $X$, $\mathbf{\Sigma_{XZ}}$ is the covariance matrix between $X$ and $Z$, and $W_{\perp}$ stands for the pseudoinverse of the matrix $W$. The $\mathbf{W}$ and $\mathbf{P}_{\mathbf{W} \mathbf{\Sigma}_{XZ}}$ defined as follows:
\begin{equation}
\label{eq:leace_w}
    \mathbf{W}=(\mathbf{\Sigma_{XX}}^{1/2})_{\perp},
\end{equation}
\begin{equation}
\label{eq:leace_p}
    \mathbf{P}_{\mathbf{W} \mathbf{\Sigma}_{XZ}}=(\mathbf{W}\mathbf{\Sigma_{XZ}})(\mathbf{W}\mathbf{\Sigma_{XZ}})_{\perp}.
\end{equation}
Then the final LEACE transformation is defined as follows:
\begin{equation}
\label{eq:leace_working}
    \hat{\mathbf{y}}(\mathbf{x}) = \mathbf{x} - \mathbf{W}_{\perp} \cdot \mathbf{P}_{\mathbf{W} \mathbf{\Sigma}_{XZ}} \cdot \mathbf{W} (\mathbf{x} - \mathbb{E}[X])
\end{equation}

\section{Fairness and UQ Oracles}
\label{sec:fairness_oracle}
In this section, we describe in detail oracle strategies for fairness and accuracy. For both strategies, we assume access to the ground-truth labels, while for fairness oracle we also use protected attributes. Accuracy oracle is built as follows -- we find all erroneously classified instances and replace predictions on these instances with ground-truth labels while keeping all other predictions unchanged. This oracle shows the best possible UQ strategy that allows the detection of all erroneous predictions and gives the maximal increase in accuracy. The same idea is behind fairness oracle, but instead of accuracy, we use fairness as a target metric. For fairness, we first replace predictions for instances, which gives the maximal increase in fairness. These predictions are chosen greedily from the erroneous ones. To measure the quality of these oracle strategies and to compare them with other scores, we calculated several metrics: FR-AUC, Acc-AUC, and FF-score-AUC. Each corresponds to the area under the target metric-rejection curve, where the target metric is fairness, accuracy, or FF-score; the area under the curve is calculated on binarized over 100 points target metric values.

\section{Datasets Statistics}
\label{sec:data_stats}

The synthetic dataset was generated as a random 2 classes classification task using \verb|make_classification| function from Scikit-learn library \cite{scikit-learn} with the following parameters: \verb|n_features=10|, \verb|n_informative=5|, \verb|n_clusters_per_class=2|, \verb|random_state=42|, \verb|n_redundant=2|. The protected attribute for the synthetic dataset is designed as a condition over the first informative feature and equals 1 if this feature is greater than 0, and 0 otherwise. The overall statistics for each dataset are presented in \Cref{tab:datasets_stats}. \Cref{tab:bios_2_distr,tab:moji_distr} shows the joint distribution of the target variable and protected attributes.

\begin{table}[t]
\centering
\resizebox{0.49\textwidth}{!}{
\begin{tabular}{lccc}
\toprule
\textbf{Dataset} & \begin{tabular}[c]{@{}l@{}}\textbf{Num. of}\\ \textbf{classes/attributes}\end{tabular} & \begin{tabular}[c]{@{}l@{}}\textbf{Protected}\\ \textbf{attribute}\end{tabular} & \textbf{Train/Val/Test}\\ \midrule
Synthetic & 2/2 & Geometric & 6k/2k/2k \\
Moji (balanced) & 2/2 & Race & 100k/8k/8k\\
Moji (imbalanced) & 2/2 & Race & 100k/5k/5k\\
Bios-2 (imbalanced) & 2/2 & Gender & 21k/3k/8k \\ 
Bios-2 (balanced) & 2/2 & Gender & 21k/1k/2k \\ 
\bottomrule
\end{tabular}
}\caption{\label{tab:datasets_stats} Dataset statistics.}\end{table}

\begin{table}[t]
\centering
\resizebox{0.4\textwidth}{!}{
\begin{tabular}{llccc}
\toprule
\multirow{2}{*}{\textbf{Split}} & \multirow{2}{*}{\textbf{Gender}} & \multicolumn{3}{c}{\textbf{Profession}}\\
\cmidrule(lr){3-5}

   &    &  \textbf{Nurse} &  \textbf{Surgeon} &   \textbf{Total}  \\
\midrule
Train & Female & 53.34 & 5.74 & 59.08 \\
 & Male & 5.50 & 35.42 & 40.92 \\
 & All & 58.84 & 41.16 & 100.00 \\\midrule
Val & Female & 53.32 & 5.08 & 58.40 \\
 & Male & 5.52 & 36.08 & 41.60 \\
 & All & 58.83 & 41.17 & 100.00 \\\midrule
Test & Female & 53.82 & 7.51 & 61.33 \\
 & Male & 5.01 & 33.66 & 38.67 \\
 & All & 58.83 & 41.17 & 100.00 \\\midrule
    Val (balanced) & Female & 26.02 & 23.98 & 50.00 \\
 & Male & 26.02 & 23.98 & 50.00 \\
 & All & 52.05 & 47.95 & 100.00 \\\midrule
Test (balanced) & Female & 20.02 & 29.98 & 50.00 \\
 & Male & 20.02 & 29.98 & 50.00 \\
 & All & 40.04 & 59.96 & 100.00 \\
\bottomrule
\end{tabular}
}\caption{\label{tab:bios_2_distr} Joint distribution for the BIOS-2 dataset.}
\end{table}
\begin{table}[h]
\centering
\resizebox{0.4\textwidth}{!}{
\begin{tabular}{llccc}
\toprule
\multirow{2}{*}{\textbf{Split}} & \multirow{2}{*}{\textbf{Ethnicity}} & \multicolumn{3}{c}{\textbf{Target}}\\
\cmidrule(lr){3-5}
     &    &  \textbf{Sad} &  \textbf{Happy} &   \textbf{Total}  \\
\midrule
Train & SA & 40.00 & 10.00 & 50.00 \\
 & AA & 10.00 & 40.00 & 50.00 \\
 & All & 50.00 & 50.00 & 100.00 \\\midrule
Val & SA & 40.02 & 9.98 & 50.00 \\
 & AA & 9.98 & 40.02 & 50.00 \\
 & All & 50.00 & 50.00 & 100.00 \\\midrule
Test & SA & 40.02 & 9.99 & 50.01 \\
 & AA & 9.99 & 40.00 & 49.99 \\
 & All & 50.01 & 49.99 & 100.00 \\\midrule
    Val (balanced) & SA & 25.00 & 25.00 & 50.00 \\
 & AA & 25.00 & 25.00 & 50.00 \\
 & All & 50.00 & 50.00 & 100.00 \\\midrule
Test (balanced) & SA & 25.01 & 25.01 & 50.01 \\
 & AA & 24.99 & 24.99 & 49.99 \\
 & All & 50.00 & 50.00 & 100.00 \\
\bottomrule
\end{tabular}
}\caption{\label{tab:moji_distr} Joint distribution for the MOJI dataset.}
\end{table}

\clearpage
\onecolumn

\section{Training Setup and Hyperparameters}
\label{sec:hyperpars}
To find an optimal set of hyperparameters, we conducted a grid search on the validation set. We used accuracy as an optimization target for standard models, and DTO for models with debiasing. The grid and optimal parameters for the standard models are described in \Cref{tab:params_vanilla}. For each debiasing method, we tuned the method's parameters and kept the training parameters of the base model  -- the grid and optimal values for debiasing methods presented in \Cref{tab:params_debiasing}. The training was conducted on a cluster with Nvidia V100 GPUs. An approximate number of GPU hours spent during the experiments is presented in \Cref{tab:gpu_hours}.

\begin{table*}[h]
\centering
\footnotesize
\resizebox{0.6\textwidth}{!}{
\begin{tabular}{llllll}
\toprule
   \textbf{Dataset} & \begin{tabular}[c]{@{}l@{}}\textbf{Num.}\\ \textbf{Epochs}\end{tabular} & \begin{tabular}[c]{@{}l@{}}\textbf{Batch}\\ \textbf{Size}\end{tabular} & \begin{tabular}[c]{@{}l@{}}\textbf{Learning}\\ \textbf{Rate}\end{tabular} & \begin{tabular}[c]{@{}l@{}}\textbf{Weight}\\ \textbf{Decay}\end{tabular} & \begin{tabular}[c]{@{}l@{}}\textbf{Dropout}\\ \textbf{Rate}\end{tabular} \\
\midrule
      MOJI imbalanced  &           20 &         32 &         1e-6 &            0 & 0.1 \\
MOJI balanced  &          20 &         32 &         1e-6 &            0 & 0.1 \\
\midrule
BIOS-2 imbalanced  &           20 &         16 &         1e-6 &            0 & 0.1 \\

BIOS-2 balanced  &          20 &         32 &         1e-6 &            1e-4 & 0.1 \\

\bottomrule
\end{tabular}
}
\caption{\label{tab:params_vanilla} Optimal training hyperparameters for BERTweet on MOJI and BERT on BIOS-2 for standard model. We use a grid search with the following grid values: batch size: [16, 32], learning rate: [1e-6, 5e-6, 1e-5, 3e-5, 5e-5], weight decay: [0, 1e-4]. The number of epochs is determined by early-stopping.}
\end{table*}
\begin{table*}[h]
\centering
\footnotesize
\resizebox{0.8\textwidth}{!}{
\begin{tabular}{llllll}
\toprule
   \textbf{Dataset} &     \begin{tabular}[c]{@{}l@{}}\textbf{Debiasing}\\ \textbf{Method}\end{tabular} & \begin{tabular}[c]{@{}l@{}}\textbf{Adv.}\\ \textbf{Lambda}\end{tabular} & \begin{tabular}[c]{@{}l@{}}\textbf{Adv. Diverse}\\ \textbf{Lambda}\end{tabular} & \begin{tabular}[c]{@{}l@{}}\textbf{INLP}\\ \textbf{by Class}\end{tabular} & \begin{tabular}[c]{@{}l@{}}\textbf{INLP Discriminator}\\ \textbf{Reweighting}\end{tabular}  \\
\midrule
      \multirow{3}{*}{Moji (imbalanced)} &   Adv &           1.0 &         - &         - &            - \\
       &      DAdv &           1.0 &         1.0 &         - &            - \\
       &      INLP &           - &         - &         False &           True \\
 \midrule
\multirow{3}{*}{Moji (balanced)} & Adv &           1.0 &         - &         - &            - \\
       &      DAdv &           1.0 &         1.0 &         - &            - \\
       &      INLP &           - &         - &         False &           False \\
\midrule
 \multirow{3}{*}{BIOS-2 (imbalanced)} & Adv &           1.0 &         - &         - &            - \\
       &      DAdv &           1.0 &         1.0 &         - &            - \\
       &      INLP &           - &         - &         False &           True \\
 \midrule
\multirow{3}{*}{BIOS-2 (balanced)} & Adv &           1.0 &         - &         - &            - \\
       &      DAdv &           1.0 &         1.0 &         - &            - \\
       &      INLP &           - &         - &         False &           True \\

\bottomrule
\end{tabular}
}
\caption{\label{tab:params_debiasing} Optimal debiasing hyperparameters for BERTweet on MOJI and BERT on BIOS-2 for various debiasing methods. The base training parameters are the same as for the vanilla model. We use a grid search with the following grid values: Adv. Lambda/Adv. Diverse Lambda: [1e-4, 1e-3, 1e-2, 1e-1, 1, 1e2, 1e3], INLP by Class/INLP Discriminator Reweighting: [False, True]. The remaining parameters for each method used default values from \cite{han-etal-2022-fairlib}. For DAdv Adv. Lambda/Adv. Diverse Lambda parameters were tuned jointly, as in \cite{han-etal-2022-fairlib}.}
\end{table*}
\begin{table*}[h]
\centering
\footnotesize
\begin{tabular}{lccc}
\toprule
   \textbf{Dataset} & \textbf{Model}  & \textbf{GPU hours} & \begin{tabular}[c]{@{}l@{}}\textbf{Num. of}\\ \textbf{Params}\end{tabular} \\
\midrule
Moji &  BERTweet &          339 &         135m \\
Bios-2 &  BERT &          119 &         110m \\
\bottomrule
\end{tabular}
\caption{\label{tab:gpu_hours} Overall computation statistics. GPU hours specify the approximate number of GPU hours spent for training and evaluating the corresponding model for all experiments on both imbalanced and balanced sets. The column Num. of Params contains the number of parameters of a single model.}
\end{table*}

\clearpage
\section{Equal Opportunity}

\label{sec:fairness_metric}
There are a numerous amount of group fairness definitions; to avoid any mismatches, we are presenting the step-by-step process of equal opportunity criterion calculation. This criterion is based on recall values, or true positive rates (TPR) for each class and protected group.

\begin{itemize}
    \item TPR (recall) for each protected group defined as follows:
    \begin{equation}
        TPR=\frac{TP}{TP+FN},
    \end{equation}
    where $TP$, $FN$ -- is true positives and false negatives for specific group.
    \item After we calculate TPR-gap:
    \begin{equation}
        \delta=\sqrt{\frac{1}{C}\sum_{c}\sum_{g}|TPR_{c,g}-\overline{TPR_{c}}|^{2}},
    \end{equation}
    here $g$ is group index, $c$ - class index, $\overline{TPR_{c}}$ - $TPR$ averaged across all groups for class $c$.
    \item Finally, we calculate fairness with the following equation:
    \begin{equation}
        Fairness=100\cdot(1-\delta).
    \end{equation}
\end{itemize}

\clearpage
\section{Additional Experiments}
\label{sec:ablation_perc}

To check how stable the proposed methods are, we compare selective debiasing results over 5\%, 10\%, and 15\% of selection for random, SR, and KL scores. The results are presented in \Cref{tab:leace_last_sel_deb,tab:leace_cls_sel_deb,tab:inlp_sel_deb}. The optimal percentage selected on the validation set from values from 1\% to 15\%; results for each dataset-method pair in \Cref{tab:opt_percentages,tab:opt_percentages_100}. In general, optimal scores are better or comparable with results on various percentages, which allows us to use this approach to detect the optimal percentage of selection.

\Cref{tab:deb_performance_100} shows the performance of selective debiasing and post-processing debiasing methods trained on a full training set. As one can see, the performance on the full set is comparable with the results on only 20\% from \Cref{tab:deb_performance}.

The results for selective debiasing with INLP trained on 20\% of data are presented in \Cref{tab:deb_performance_inlp}. INLP-based selective debiasing improves the FF-score only on MOJI-balanced, while on other datasets, it is consistent with the base inference-time debiasing method. INLP-based approaches overall fall behind the corresponding LEACE-based techniques.

\begin{table*}[h] 
\centering
\resizebox{1.\textwidth}{!}{\begin{tabular}{ll||c|cc|cc||ccc|ccc|ccc}
\toprule
\multicolumn{2}{c||}{\textbf{Debiasing method type}} & \textbf{No debiasing} & \multicolumn{2}{|c|}{\textbf{At-training}} & \multicolumn{2}{c||}{\textbf{Pre-processing}} & \multicolumn{9}{c}{\textbf{Post-processing \& Selective}} \\
\toprule
\textbf{Dataset} & \textbf{Metric} & \textbf{Standard} & \textbf{Adv} & \textbf{DAdv} & \textbf{BTEO} & \textbf{BTJ} & \begin{tabular}[c]{@{}l@{}}\textbf{LEACE-}\\ \textbf{last}\\ \end{tabular} & \begin{tabular}[c]{@{}l@{}}\textbf{LEACE-}\\ \textbf{last+SR,}\\ \textbf{opt.} \textbf{perc.}\end{tabular} & \begin{tabular}[c]{@{}l@{}}\textbf{LEACE-}\\ \textbf{last+KL,}\\ \textbf{opt.} \textbf{perc.}\end{tabular} & \begin{tabular}[c]{@{}l@{}}\textbf{LEACE-}\\ \textbf{cls}\\ \end{tabular} & \begin{tabular}[c]{@{}l@{}}\textbf{LEACE-}\\ \textbf{cls+SR,}\\ \textbf{opt.} \textbf{perc.}\end{tabular} & \begin{tabular}[c]{@{}l@{}}\textbf{LEACE-}\\ \textbf{cls+KL,}\\ \textbf{opt.} \textbf{perc.}\end{tabular} & \textbf{INLP} & \begin{tabular}[c]{@{}l@{}}\textbf{INLP+SR,}\\ \textbf{opt.}  \textbf{perc.}\end{tabular} & \begin{tabular}[c]{@{}l@{}}\textbf{INLP+KL,}\\ \textbf{opt.}  \textbf{perc.}\end{tabular} \\
\midrule
\multirow[c]{4}{*}{\begin{tabular}[c]{@{}l@{}}MOJI\\ imbalanced\\ \end{tabular}} & Fairness $\uparrow$ & 61.8\tiny{$\pm$0.7} & 73.7\tiny{$\pm$0.6} & 73.4\tiny{$\pm$0.4} & \textbf{75.2\tiny{$\pm$0.6}} & 74.8\tiny{$\pm$0.6} & 75.7\tiny{$\pm$2.6} & 68.5\tiny{$\pm$1.3} & \textbf{75.9\tiny{$\pm$1.3}} & 74.5\tiny{$\pm$2.4} & 68.4\tiny{$\pm$1.2} & \textbf{77.0\tiny{$\pm$0.9}} & \underline{\textbf{88.2\tiny{$\pm$6.3}}} & 64.1\tiny{$\pm$1.7} & 73.2\tiny{$\pm$1.3} \\
 & Accuracy $\uparrow$ & \underline{\textbf{79.1\tiny{$\pm$0.7}}} & 72.0\tiny{$\pm$0.7} & 72.4\tiny{$\pm$0.5} & 73.6\tiny{$\pm$0.6} & 73.2\tiny{$\pm$0.4} & 68.3\tiny{$\pm$2.3} & \textbf{77.7\tiny{$\pm$1.0}} & 72.2\tiny{$\pm$1.1} & 66.8\tiny{$\pm$2.5} & \textbf{77.6\tiny{$\pm$1.0}} & 71.6\tiny{$\pm$1.1} & 59.9\tiny{$\pm$7.3} & \textbf{77.6\tiny{$\pm$1.3}} & 71.6\tiny{$\pm$1.9} \\
 & DTO $\downarrow$ & 43.6\tiny{$\pm$0.6} & 38.4\tiny{$\pm$0.5} & 38.3\tiny{$\pm$0.4} & \underline{\textbf{36.2\tiny{$\pm$0.1}}} & 36.7\tiny{$\pm$0.4} & 40.0\tiny{$\pm$3.4} & 38.6\tiny{$\pm$0.7} & \textbf{36.8\tiny{$\pm$0.7}} & 41.9\tiny{$\pm$3.4} & 38.8\tiny{$\pm$0.7} & \textbf{36.6\tiny{$\pm$1.0}} & 42.6\tiny{$\pm$5.1} & 42.4\tiny{$\pm$1.3} & \textbf{39.0\tiny{$\pm$1.8}} \\
 & FF-score $\uparrow$ & 69.4\tiny{$\pm$0.4} & 72.8\tiny{$\pm$0.4} & 72.9\tiny{$\pm$0.3} & \underline{\textbf{74.4\tiny{$\pm$0.1}}} & 74.0\tiny{$\pm$0.3} & 71.8\tiny{$\pm$2.4} & 72.8\tiny{$\pm$0.5} & \textbf{74.0\tiny{$\pm$0.5}} & 70.4\tiny{$\pm$2.4} & 72.7\tiny{$\pm$0.5} & \textbf{74.2\tiny{$\pm$0.7}} & 70.8\tiny{$\pm$3.4} & 70.2\tiny{$\pm$0.9} & \textbf{72.4\tiny{$\pm$1.3}} \\\hline
\multirow[c]{4}{*}{\begin{tabular}[c]{@{}l@{}}MOJI\\ balanced\\ \end{tabular}} & Fairness $\uparrow$ & 69.5\tiny{$\pm$0.2} & 83.8\tiny{$\pm$0.8} & 84.7\tiny{$\pm$1.5} & 85.5\tiny{$\pm$0.5} & \textbf{85.6\tiny{$\pm$0.6}} & 79.7\tiny{$\pm$3.5} & 77.0\tiny{$\pm$0.9} & \textbf{86.7\tiny{$\pm$0.6}} & 77.0\tiny{$\pm$3.4} & 77.0\tiny{$\pm$0.8} & \underline{\textbf{87.3\tiny{$\pm$0.7}}} & \textbf{77.3\tiny{$\pm$5.6}} & 70.4\tiny{$\pm$1.3} & 74.0\tiny{$\pm$2.8} \\
 & Accuracy $\uparrow$ & 71.9\tiny{$\pm$0.4} & 74.0\tiny{$\pm$0.4} & 74.1\tiny{$\pm$0.6} & \underline{\textbf{74.8\tiny{$\pm$0.3}}} & 74.5\tiny{$\pm$0.4} & 73.6\tiny{$\pm$0.7} & \textbf{74.0\tiny{$\pm$0.4}} & 73.9\tiny{$\pm$0.2} & 73.0\tiny{$\pm$0.9} & \textbf{74.0\tiny{$\pm$0.4}} & 73.6\tiny{$\pm$0.5} & 65.9\tiny{$\pm$4.6} & \textbf{71.8\tiny{$\pm$0.4}} & 69.0\tiny{$\pm$1.8} \\
 & DTO $\downarrow$ & 41.5\tiny{$\pm$0.4} & 30.7\tiny{$\pm$0.7} & 30.1\tiny{$\pm$0.7} & \underline{\textbf{29.0\tiny{$\pm$0.1}}} & 29.3\tiny{$\pm$0.4} & 33.4\tiny{$\pm$2.7} & 34.7\tiny{$\pm$0.7} & \textbf{29.3\tiny{$\pm$0.4}} & 35.5\tiny{$\pm$3.0} & 34.7\tiny{$\pm$0.6} & \textbf{29.3\tiny{$\pm$0.5}} & 41.3\tiny{$\pm$4.5} & 40.9\tiny{$\pm$0.9} & \textcolor{gray}{\textbf{40.6\tiny{$\pm$2.3}}} \\
 & FF-score $\uparrow$ & 70.7\tiny{$\pm$0.3} & 78.6\tiny{$\pm$0.5} & 79.1\tiny{$\pm$0.6} & \textbf{79.8\tiny{$\pm$0.1}} & 79.6\tiny{$\pm$0.3} & 76.5\tiny{$\pm$2.0} & 75.5\tiny{$\pm$0.5} & \textbf{79.8\tiny{$\pm$0.3}} & 74.9\tiny{$\pm$2.1} & 75.5\tiny{$\pm$0.4} & \underline{\textbf{79.9\tiny{$\pm$0.4}}} & 71.0\tiny{$\pm$3.4} & 71.1\tiny{$\pm$0.7} & \textcolor{gray}{\textbf{71.4\tiny{$\pm$1.6}}} \\\hline
\multirow[c]{4}{*}{\begin{tabular}[c]{@{}l@{}}BIOS-2\\ imbalanced\\ \end{tabular}} & Fairness $\uparrow$ & 90.4\tiny{$\pm$0.8} & \underline{\textbf{97.2\tiny{$\pm$0.8}}} & 96.4\tiny{$\pm$0.4} & 95.8\tiny{$\pm$1.0} & 96.6\tiny{$\pm$0.8} & \textbf{93.3\tiny{$\pm$8.1}} & 93.1\tiny{$\pm$2.3} & \textcolor{gray}{92.9\tiny{$\pm$2.1}} & 78.0\tiny{$\pm$5.5} & 95.2\tiny{$\pm$2.5} & \textbf{96.4\tiny{$\pm$1.0}} & 91.6\tiny{$\pm$1.6} & \textbf{91.9\tiny{$\pm$0.8}} & \textcolor{gray}{91.5\tiny{$\pm$1.5}} \\
 & Accuracy $\uparrow$ & \underline{\textbf{96.7\tiny{$\pm$0.1}}} & 94.8\tiny{$\pm$0.4} & 95.0\tiny{$\pm$0.3} & 95.2\tiny{$\pm$0.3} & 95.0\tiny{$\pm$0.5} & 61.1\tiny{$\pm$4.0} & 94.6\tiny{$\pm$0.3} & \textbf{94.7\tiny{$\pm$0.2}} & 65.4\tiny{$\pm$5.6} & \textbf{94.6\tiny{$\pm$0.1}} & 93.2\tiny{$\pm$0.3} & \textbf{95.9\tiny{$\pm$0.8}} & \textbf{95.9\tiny{$\pm$0.6}} & \textcolor{gray}{\textbf{95.9\tiny{$\pm$0.8}}} \\
 & DTO $\downarrow$ & 10.1\tiny{$\pm$0.7} & \underline{\textbf{5.9\tiny{$\pm$0.2}}} & 6.2\tiny{$\pm$0.2} & 6.5\tiny{$\pm$0.6} & 6.1\tiny{$\pm$0.3} & 40.4\tiny{$\pm$2.2} & \textbf{8.9\tiny{$\pm$1.7}} & \textcolor{gray}{9.0\tiny{$\pm$1.6}} & 41.7\tiny{$\pm$2.0} & \textbf{7.4\tiny{$\pm$1.8}} & 7.7\tiny{$\pm$0.6} & 9.5\tiny{$\pm$1.1} & \textbf{9.1\tiny{$\pm$0.4}} & \textcolor{gray}{9.5\tiny{$\pm$1.1}} \\
 & FF-score $\uparrow$ & 93.5\tiny{$\pm$0.4} & \underline{\textbf{96.0\tiny{$\pm$0.2}}} & 95.7\tiny{$\pm$0.1} & 95.5\tiny{$\pm$0.4} & 95.8\tiny{$\pm$0.2} & 73.5\tiny{$\pm$2.0} & \textbf{93.8\tiny{$\pm$1.2}} & \textcolor{gray}{93.7\tiny{$\pm$1.1}} & 70.7\tiny{$\pm$1.2} & \textbf{94.9\tiny{$\pm$1.3}} & 94.8\tiny{$\pm$0.6} & 93.7\tiny{$\pm$0.6} & \textbf{93.8\tiny{$\pm$0.2}} & \textcolor{gray}{93.6\tiny{$\pm$0.6}} \\\hline
\multirow[c]{4}{*}{\begin{tabular}[c]{@{}l@{}}BIOS-2\\ balanced\\ \end{tabular}} & Fairness $\uparrow$ & 89.7\tiny{$\pm$0.6} & 97.8\tiny{$\pm$0.8} & \underline{\textbf{98.0\tiny{$\pm$0.8}}} & 95.9\tiny{$\pm$0.8} & 96.4\tiny{$\pm$0.3} & 91.2\tiny{$\pm$10.2} & 93.2\tiny{$\pm$2.6} & \textbf{94.3\tiny{$\pm$3.6}} & 74.7\tiny{$\pm$9.2} & 96.7\tiny{$\pm$1.6} & \textbf{97.6\tiny{$\pm$1.1}} & \textbf{91.8\tiny{$\pm$1.1}} & 91.4\tiny{$\pm$0.9} & \textbf{91.8\tiny{$\pm$1.1}} \\
 & Accuracy $\uparrow$ & 92.4\tiny{$\pm$0.3} & 91.9\tiny{$\pm$0.6} & 91.9\tiny{$\pm$1.5} & 92.6\tiny{$\pm$0.5} & \underline{\textbf{92.9\tiny{$\pm$0.6}}} & 50.4\tiny{$\pm$8.8} & \textbf{90.7\tiny{$\pm$1.2}} & 90.0\tiny{$\pm$1.8} & 64.0\tiny{$\pm$9.7} & \textbf{91.9\tiny{$\pm$0.9}} & 90.6\tiny{$\pm$1.4} & 90.7\tiny{$\pm$1.2} & \textbf{91.1\tiny{$\pm$1.1}} & 90.7\tiny{$\pm$1.1} \\
 & DTO $\downarrow$ & 12.8\tiny{$\pm$0.6} & 8.5\tiny{$\pm$0.4} & 8.4\tiny{$\pm$1.4} & 8.5\tiny{$\pm$0.2} & \underline{\textbf{8.0\tiny{$\pm$0.6}}} & 51.9\tiny{$\pm$5.1} & \textbf{11.6\tiny{$\pm$2.4}} & \textcolor{gray}{11.7\tiny{$\pm$3.3}} & 45.8\tiny{$\pm$3.5} & \textbf{8.8\tiny{$\pm$1.4}} & 9.7\tiny{$\pm$1.6} & 12.5\tiny{$\pm$1.2} & \textbf{12.4\tiny{$\pm$1.1}} & \textcolor{gray}{\textbf{12.4\tiny{$\pm$1.1}}} \\
 & FF-score $\uparrow$ & 91.1\tiny{$\pm$0.4} & 94.7\tiny{$\pm$0.1} & \underline{\textbf{94.9\tiny{$\pm$0.7}}} & 94.2\tiny{$\pm$0.2} & 94.6\tiny{$\pm$0.3} & 63.7\tiny{$\pm$3.6} & 91.9\tiny{$\pm$1.9} & \textcolor{gray}{\textbf{92.1\tiny{$\pm$2.6}}} & 67.7\tiny{$\pm$2.4} & \textbf{94.2\tiny{$\pm$1.2}} & 94.0\tiny{$\pm$1.1} & 91.2\tiny{$\pm$0.9} & \textbf{91.3\tiny{$\pm$0.8}} & \textcolor{gray}{\textbf{91.3\tiny{$\pm$0.8}}} \\

 \bottomrule
\end{tabular}
}\caption{\label{tab:deb_performance_100} Comparison of debiasing methods and selective debiasing; the post-processing methods trained on full training set. The best results in the group are in bold, and the best results overall are underlined. The gray color corresponds to the results with p-value > 0.05 with respect to standard model.}\end{table*}

\begin{table*}[ht] \resizebox{\textwidth}{!}{\begin{tabular}{lllllllllllllll}
\toprule
 \textbf{Dataset} & \textbf{Standard} & \textbf{LEACE} & \begin{tabular}[c]{@{}l@{}}\textbf{Random,}\\ \textbf{5\%}\end{tabular} & \begin{tabular}[c]{@{}l@{}}\textbf{SR,}\\ \textbf{5\%}\end{tabular} & \begin{tabular}[c]{@{}l@{}}\textbf{KL,}\\ \textbf{5\%}\end{tabular} & \begin{tabular}[c]{@{}l@{}}\textbf{Random,}\\ \textbf{10\%}\end{tabular} & \begin{tabular}[c]{@{}l@{}}\textbf{SR,}\\ \textbf{10\%}\end{tabular} & \begin{tabular}[c]{@{}l@{}}\textbf{KL,}\\ \textbf{10\%}\end{tabular} & \begin{tabular}[c]{@{}l@{}}\textbf{Random,}\\ \textbf{15\%}\end{tabular} & \begin{tabular}[c]{@{}l@{}}\textbf{SR,}\\ \textbf{15\%}\end{tabular} & \begin{tabular}[c]{@{}l@{}}\textbf{KL,}\\ \textbf{15\%}\end{tabular} & \begin{tabular}[c]{@{}l@{}}\textbf{Random,}\\ \textbf{optimal}\\ \textbf{percentage}\end{tabular} & \begin{tabular}[c]{@{}l@{}}\textbf{SR,}\\ \textbf{optimal}\\ \textbf{percentage}\end{tabular} & \begin{tabular}[c]{@{}l@{}}\textbf{KL,}\\ \textbf{optimal}\\ \textbf{percentage}\end{tabular} \\
 \midrule
MOJI imbalanced & 69.4\tiny{$\pm$0.4} & 71.8\tiny{$\pm$2.6} & 70.6\tiny{$\pm$0.3} & 70.8\tiny{$\pm$0.5} & 72.5\tiny{$\pm$0.6} & 71.4\tiny{$\pm$0.3} & 71.8\tiny{$\pm$0.6} & 73.7\tiny{$\pm$0.7} & 72.0\tiny{$\pm$0.4} & 72.8\tiny{$\pm$0.5} & \textbf{74.1\tiny{$\pm$0.3}} & 72.0\tiny{$\pm$0.4} & 72.8\tiny{$\pm$0.5} & \textbf{74.1\tiny{$\pm$0.3}} \\
MOJI balanced & 70.7\tiny{$\pm$0.3} & 76.5\tiny{$\pm$2.2} & 71.8\tiny{$\pm$0.2} & 72.4\tiny{$\pm$0.1} & 75.4\tiny{$\pm$0.2} & 72.5\tiny{$\pm$0.0} & 74.0\tiny{$\pm$0.2} & 78.2\tiny{$\pm$0.3} & 73.3\tiny{$\pm$0.1} & 75.5\tiny{$\pm$0.5} & \textbf{79.8\tiny{$\pm$0.2}} & 73.3\tiny{$\pm$0.1} & 75.5\tiny{$\pm$0.5} & \textbf{79.8\tiny{$\pm$0.2}} \\
BIOS-2 imbalanced & 93.5\tiny{$\pm$0.4} & 72.8\tiny{$\pm$2.3} & 92.9\tiny{$\pm$0.5} & 93.7\tiny{$\pm$1.0} & 93.3\tiny{$\pm$2.0} & 92.0\tiny{$\pm$0.8} & 93.7\tiny{$\pm$1.3} & 90.3\tiny{$\pm$2.1} & 91.1\tiny{$\pm$1.2} & 93.1\tiny{$\pm$1.1} & 86.3\tiny{$\pm$2.2} & 93.4\tiny{$\pm$0.4} & \textbf{93.8\tiny{$\pm$1.2}} & 93.2\tiny{$\pm$2.3} \\
BIOS-2 balanced & 91.1\tiny{$\pm$0.4} & 63.0\tiny{$\pm$4.6} & 90.6\tiny{$\pm$0.3} & 91.6\tiny{$\pm$1.0} & 92.0\tiny{$\pm$2.1} & 89.3\tiny{$\pm$0.9} & 92.2\tiny{$\pm$1.9} & 89.6\tiny{$\pm$2.4} & 88.7\tiny{$\pm$1.2} & \textbf{92.6\tiny{$\pm$2.2}} & 85.4\tiny{$\pm$2.4} & 90.9\tiny{$\pm$0.3} & 92.3\tiny{$\pm$1.9} & 91.9\tiny{$\pm$2.9} \\
\bottomrule
\end{tabular}
}\caption{\label{tab:leace_last_sel_deb} FF-score of selective debiasing for LEACE on the last layer for various percentages.}\end{table*}

\begin{table*}[ht] \resizebox{\textwidth}{!}{\begin{tabular}{lllllllllllllll}
\toprule
 \textbf{Dataset} & \textbf{Standard} & \textbf{LEACE} & \begin{tabular}[c]{@{}l@{}}\textbf{Random,}\\ \textbf{5\%}\end{tabular} & \begin{tabular}[c]{@{}l@{}}\textbf{SR,}\\ \textbf{5\%}\end{tabular} & \begin{tabular}[c]{@{}l@{}}\textbf{KL,}\\ \textbf{5\%}\end{tabular} & \begin{tabular}[c]{@{}l@{}}\textbf{Random,}\\ \textbf{10\%}\end{tabular} & \begin{tabular}[c]{@{}l@{}}\textbf{SR,}\\ \textbf{10\%}\end{tabular} & \begin{tabular}[c]{@{}l@{}}\textbf{KL,}\\ \textbf{10\%}\end{tabular} & \begin{tabular}[c]{@{}l@{}}\textbf{Random,}\\ \textbf{15\%}\end{tabular} & \begin{tabular}[c]{@{}l@{}}\textbf{SR,}\\ \textbf{15\%}\end{tabular} & \begin{tabular}[c]{@{}l@{}}\textbf{KL,}\\ \textbf{15\%}\end{tabular} & \begin{tabular}[c]{@{}l@{}}\textbf{Random,}\\ \textbf{optimal}\\ \textbf{percentage}\end{tabular} & \begin{tabular}[c]{@{}l@{}}\textbf{SR,}\\ \textbf{optimal}\\ \textbf{percentage}\end{tabular} & \begin{tabular}[c]{@{}l@{}}\textbf{KL,}\\ \textbf{optimal}\\ \textbf{percentage}\end{tabular} \\
 \midrule
MOJI imbalanced & 69.4\tiny{$\pm$0.4} & 70.8\tiny{$\pm$3.0} & 70.7\tiny{$\pm$0.4} & 70.8\tiny{$\pm$0.5} & 72.5\tiny{$\pm$0.8} & 71.5\tiny{$\pm$0.4} & 71.7\tiny{$\pm$0.5} & 74.2\tiny{$\pm$0.6} & 72.1\tiny{$\pm$0.5} & 72.7\tiny{$\pm$0.4} & \textbf{74.4\tiny{$\pm$0.8}} & 72.1\tiny{$\pm$0.5} & 72.7\tiny{$\pm$0.4} & \textbf{74.4\tiny{$\pm$0.8}} \\
MOJI balanced & 70.7\tiny{$\pm$0.3} & 75.2\tiny{$\pm$2.6} & 71.8\tiny{$\pm$0.3} & 72.4\tiny{$\pm$0.1} & 75.5\tiny{$\pm$0.2} & 72.7\tiny{$\pm$0.2} & 74.0\tiny{$\pm$0.2} & 78.7\tiny{$\pm$0.3} & 73.5\tiny{$\pm$0.1} & 75.5\tiny{$\pm$0.4} & \textbf{80.0\tiny{$\pm$0.4}} & 73.5\tiny{$\pm$0.1} & 75.5\tiny{$\pm$0.4} & \textbf{80.0\tiny{$\pm$0.4}} \\
BIOS-2 imbalanced & 93.5\tiny{$\pm$0.4} & 69.6\tiny{$\pm$1.7} & 93.4\tiny{$\pm$0.7} & 94.6\tiny{$\pm$1.0} & \textbf{95.1\tiny{$\pm$0.4}} & 93.1\tiny{$\pm$1.0} & 94.7\tiny{$\pm$0.9} & 91.0\tiny{$\pm$1.0} & 92.9\tiny{$\pm$1.3} & 93.2\tiny{$\pm$0.6} & 85.1\tiny{$\pm$1.1} & 93.5\tiny{$\pm$0.4} & 94.7\tiny{$\pm$1.2} & 94.9\tiny{$\pm$0.4} \\
BIOS-2 balanced & 91.1\tiny{$\pm$0.4} & 67.5\tiny{$\pm$3.0} & 91.2\tiny{$\pm$0.5} & 92.4\tiny{$\pm$0.6} & 93.9\tiny{$\pm$1.1} & 90.8\tiny{$\pm$1.1} & 93.7\tiny{$\pm$1.2} & 90.1\tiny{$\pm$2.3} & 90.8\tiny{$\pm$1.3} & \textbf{94.4\tiny{$\pm$0.9}} & 83.4\tiny{$\pm$2.2} & 91.2\tiny{$\pm$0.7} & 94.2\tiny{$\pm$1.2} & 93.9\tiny{$\pm$1.1} \\
\bottomrule
\end{tabular}
}\caption{\label{tab:leace_cls_sel_deb} FF-score of selective debiasing for LEACE on the classifier level for various percentages.}\end{table*}

\begin{table*}[h] \resizebox{\textwidth}{!}{\begin{tabular}{lllllllllllllll}
\toprule
 \textbf{Dataset} & \textbf{Standard} & \textbf{INLP} & \begin{tabular}[c]{@{}l@{}}\textbf{Random,}\\ \textbf{5\%}\end{tabular} & \begin{tabular}[c]{@{}l@{}}\textbf{SR,}\\ \textbf{5\%}\end{tabular} & \begin{tabular}[c]{@{}l@{}}\textbf{KL,}\\ \textbf{5\%}\end{tabular} & \begin{tabular}[c]{@{}l@{}}\textbf{Random,}\\ \textbf{10\%}\end{tabular} & \begin{tabular}[c]{@{}l@{}}\textbf{SR,}\\ \textbf{10\%}\end{tabular} & \begin{tabular}[c]{@{}l@{}}\textbf{KL,}\\ \textbf{10\%}\end{tabular} & \begin{tabular}[c]{@{}l@{}}\textbf{Random,}\\ \textbf{15\%}\end{tabular} & \begin{tabular}[c]{@{}l@{}}\textbf{SR,}\\ \textbf{15\%}\end{tabular} & \begin{tabular}[c]{@{}l@{}}\textbf{KL,}\\ \textbf{15\%}\end{tabular} & \begin{tabular}[c]{@{}l@{}}\textbf{Random,}\\ \textbf{optimal}\\ \textbf{percentage}\end{tabular} & \begin{tabular}[c]{@{}l@{}}\textbf{SR,}\\ \textbf{optimal}\\ \textbf{percentage}\end{tabular} & \begin{tabular}[c]{@{}l@{}}\textbf{KL,}\\ \textbf{optimal}\\ \textbf{percentage}\end{tabular} \\
  \midrule
  
MOJI imbalanced & 69.4\tiny{$\pm$0.4} & \textbf{71.9\tiny{$\pm$2.2}} & 70.0\tiny{$\pm$0.4} & 69.5\tiny{$\pm$0.3} & 71.0\tiny{$\pm$0.8} & 70.1\tiny{$\pm$0.4} & 69.6\tiny{$\pm$0.5} & 71.8\tiny{$\pm$1.0} & 70.2\tiny{$\pm$0.5} & 70.1\tiny{$\pm$0.5} & \textbf{71.9\tiny{$\pm$1.3}} & 70.2\tiny{$\pm$0.5} & 70.0\tiny{$\pm$0.6} & \textbf{71.9\tiny{$\pm$1.3}} \\
MOJI balanced & 70.7\tiny{$\pm$0.3} & 71.9\tiny{$\pm$3.2} & 71.0\tiny{$\pm$0.4} & 71.2\tiny{$\pm$0.1} & 72.0\tiny{$\pm$0.6} & 71.2\tiny{$\pm$0.4} & 71.6\tiny{$\pm$0.4} & 72.5\tiny{$\pm$0.9} & 71.3\tiny{$\pm$0.5} & 71.8\tiny{$\pm$0.5} & \textbf{72.8\tiny{$\pm$1.0}} & 71.3\tiny{$\pm$0.5} & 71.8\tiny{$\pm$0.4} & \textbf{72.8\tiny{$\pm$1.0}} \\
BIOS-2 imbalanced & 93.5\tiny{$\pm$0.4} & 93.8\tiny{$\pm$0.7} & 93.5\tiny{$\pm$0.4} & \textbf{93.9\tiny{$\pm$0.7}} & 93.8\tiny{$\pm$0.8} & 93.5\tiny{$\pm$0.4} & 93.8\tiny{$\pm$0.7} & 93.8\tiny{$\pm$0.8} & 93.6\tiny{$\pm$0.4} & 93.8\tiny{$\pm$0.7} & 93.8\tiny{$\pm$0.8} & 93.6\tiny{$\pm$0.4} & \textbf{93.9\tiny{$\pm$0.7}} & 93.8\tiny{$\pm$0.8} \\
BIOS-2 balanced & 91.1\tiny{$\pm$0.4} & 91.8\tiny{$\pm$1.1} & 91.1\tiny{$\pm$0.4} & 91.5\tiny{$\pm$0.9} & 91.6\tiny{$\pm$1.1} & 91.1\tiny{$\pm$0.5} & 91.8\tiny{$\pm$1.1} & 91.7\tiny{$\pm$1.1} & 91.2\tiny{$\pm$0.5} & \textbf{91.9\tiny{$\pm$1.2}} & 91.7\tiny{$\pm$1.2} & 91.2\tiny{$\pm$0.5} & \textbf{91.9\tiny{$\pm$1.2}} & 91.6\tiny{$\pm$1.1} \\
\bottomrule
\end{tabular}
}\caption{\label{tab:inlp_sel_deb} FF-score of selective debiasing for INLP for various percentages.}\end{table*}

\begin{table*}[h] 
\footnotesize
\centering\resizebox{0.6\textwidth}{!}{\begin{tabular}{l|lll|lll|lll}

\toprule
 \multirow{2}{*}{\textbf{Dataset}} & \multicolumn{3}{c|}{\textbf{LEACE-last}} & \multicolumn{3}{c|}{\textbf{LEACE-cls}} & \multicolumn{3}{c}{\textbf{INLP}} \\
  & Random & SR & KL & Random & SR & KL & Random & SR & KL \\
 \midrule
MOJI imbalanced & 15 & 15 & 14   & 15 & 15 & 15    & 15 & 14 & 15\\
MOJI balanced & 15 & 15 & 15   & 15 & 15 & 15   & 15 & 13 & 15 \\
BIOS-2 imbalanced & 1 & 6 & 6    & 1 & 6 & 4    & 15 & 6 & 14 \\
BIOS-2 balanced & 1 & 11 & 7   & 7 & 12 & 5    & 8 & 15 & 5 \\
\bottomrule
\end{tabular}
}\caption{\label{tab:opt_percentages} Optimal selection percentages for various debiasing methods.}\end{table*}
\begin{table*}[h] 
\footnotesize
\centering\resizebox{0.6\textwidth}{!}{\begin{tabular}{l|lll|lll|lll}

\toprule
 \multirow{2}{*}{\textbf{Dataset}} & \multicolumn{3}{c|}{\textbf{LEACE-last}} & \multicolumn{3}{c|}{\textbf{LEACE-cls}} & \multicolumn{3}{c}{\textbf{INLP}} \\
  & Random & SR & KL & Random & SR & KL & Random & SR & KL \\
 \midrule
MOJI imbalanced & 15 & 15 & 15   & 15 & 15 & 15    & 15 & 15 & 14\\
MOJI balanced & 15 & 15 & 15   & 15 & 15 & 15   & 15 & 13 & 11 \\
BIOS-2 imbalanced & 1 & 6 & 3    & 1 & 6 & 4    & 9 & 12 & 8 \\
BIOS-2 balanced & 1 & 11 & 6   & 7 & 12 & 5    & 8 & 15 & 13 \\
\bottomrule
\end{tabular}
}\caption{\label{tab:opt_percentages_100} Optimal selection percentages for various debiasing methods, the post-processing methods trained on full training set.}\end{table*}
\begin{table*}[h] 
\centering
\resizebox{0.75\textwidth}{!}{\begin{tabular}{ll||c|cc|cc||ccc}
\toprule
\multicolumn{2}{c||}{\textbf{Debiasing method type}} & \textbf{No debiasing} & \multicolumn{2}{c|}{\textbf{At-training}} & \multicolumn{2}{c||}{\textbf{Pre-processing}} & \multicolumn{3}{c}{\textbf{Post-processing \& Selective}} \\
\toprule
\textbf{Dataset} & \textbf{Metric} & \textbf{Standard} & \textbf{Adv} & \textbf{DAdv} & \textbf{BTEO} & \textbf{BTJ} & \textbf{INLP} & \begin{tabular}[c]{@{}l@{}}\textbf{INLP+SR,}\\ \textbf{opt.}  \textbf{perc.}\end{tabular} & \begin{tabular}[c]{@{}l@{}}\textbf{INLP+KL,}\\ \textbf{opt.}  \textbf{perc.}\end{tabular} \\
\midrule
\multirow[c]{4}{*}{\begin{tabular}[c]{@{}l@{}}MOJI\\ imbalanced\\ \end{tabular}} & Fairness $\uparrow$ & 61.8\tiny{$\pm$0.7} & 73.7\tiny{$\pm$0.6} & 73.4\tiny{$\pm$0.4} & \textbf{75.2\tiny{$\pm$0.6}} & 74.8\tiny{$\pm$0.6} & \underline{\textbf{77.3\tiny{$\pm$7.3}}} & 63.5\tiny{$\pm$1.1} & 70.5\tiny{$\pm$2.4} \\
 & Accuracy $\uparrow$ & \underline{\textbf{79.1\tiny{$\pm$0.7}}} & 72.0\tiny{$\pm$0.7} & 72.4\tiny{$\pm$0.5} & 73.6\tiny{$\pm$0.6} & 73.2\tiny{$\pm$0.4} & 68.4\tiny{$\pm$6.8} & \textbf{78.0\tiny{$\pm$0.6}} & 73.5\tiny{$\pm$1.4} \\
 & DTO $\downarrow$ & 43.6\tiny{$\pm$0.6} & 38.4\tiny{$\pm$0.5} & 38.3\tiny{$\pm$0.4} & \underline{\textbf{36.2\tiny{$\pm$0.1}}} & 36.7\tiny{$\pm$0.4}  & 40.0\tiny{$\pm$3.5} & 42.6\tiny{$\pm$0.8} & \textbf{39.7\tiny{$\pm$1.8}} \\
 & FF-score $\uparrow$ & 69.4\tiny{$\pm$0.4} & 72.8\tiny{$\pm$0.4} & 72.9\tiny{$\pm$0.3} & \underline{\textbf{74.4\tiny{$\pm$0.1}}} & \textbf{74.0\tiny{$\pm$0.3}} & \textbf{71.9\tiny{$\pm$2.2}} & 70.0\tiny{$\pm$0.6} & \textbf{71.9\tiny{$\pm$1.3}} \\
\midrule
\multirow[c]{4}{*}{\begin{tabular}[c]{@{}l@{}}MOJI\\ balanced\\ \end{tabular}} & Fairness $\uparrow$ & 69.5\tiny{$\pm$0.2} & 83.8\tiny{$\pm$0.8} & 84.7\tiny{$\pm$1.5} & 85.5\tiny{$\pm$0.5} & \textbf{85.6\tiny{$\pm$0.6}}  & \underline{\textbf{85.8\tiny{$\pm$8.3}}} & 71.7\tiny{$\pm$0.6} & 77.9\tiny{$\pm$4.4} \\
 & Accuracy $\uparrow$ & 71.9\tiny{$\pm$0.4} & 74.0\tiny{$\pm$0.4} & 74.1\tiny{$\pm$0.6} & \underline{\textbf{74.8\tiny{$\pm$0.3}}} & 74.5\tiny{$\pm$0.4}  & 63.0\tiny{$\pm$6.9} & \textbf{71.9\tiny{$\pm$0.4}} & 68.5\tiny{$\pm$2.0} \\
 & DTO $\downarrow$ & 41.5\tiny{$\pm$0.4} & 30.7\tiny{$\pm$0.7} & 30.1\tiny{$\pm$0.7} & \underline{\textbf{29.0\tiny{$\pm$0.1}}} & 29.3\tiny{$\pm$0.4} & 40.9\tiny{$\pm$4.4} & 39.9\tiny{$\pm$0.6} & \textbf{38.8\tiny{$\pm$1.2}} \\
 & FF-score $\uparrow$ & 70.7\tiny{$\pm$0.3} & 78.6\tiny{$\pm$0.5} & 79.1\tiny{$\pm$0.6} & \textbf{79.8\tiny{$\pm$0.1}} & 79.6\tiny{$\pm$0.3} & 71.9\tiny{$\pm$3.2} & 71.8\tiny{$\pm$0.4} & \textbf{72.8\tiny{$\pm$1.0}} \\
\midrule
\multirow[c]{4}{*}{\begin{tabular}[c]{@{}l@{}}BIOS-2\\ imbalanced\\ \end{tabular}} & Fairness $\uparrow$ & 90.4\tiny{$\pm$0.8} & \underline{\textbf{97.2\tiny{$\pm$0.8}}} & 96.4\tiny{$\pm$0.4} & 95.8\tiny{$\pm$1.0} & 96.6\tiny{$\pm$0.8} & \textbf{92.0\tiny{$\pm$1.6}} & 91.7\tiny{$\pm$1.4} & \textcolor{gray}{91.9\tiny{$\pm$1.7}} \\
 & Accuracy $\uparrow$ & \underline{\textbf{96.7\tiny{$\pm$0.1}}} & 94.8\tiny{$\pm$0.4} & 95.0\tiny{$\pm$0.3} & 95.2\tiny{$\pm$0.3} & 95.0\tiny{$\pm$0.5} & 95.8\tiny{$\pm$0.6} & \textbf{96.2\tiny{$\pm$0.3}} & 95.8\tiny{$\pm$0.6} \\
 & DTO $\downarrow$ & 10.1\tiny{$\pm$0.7} & \underline{\textbf{5.9\tiny{$\pm$0.2}}} & 6.2\tiny{$\pm$0.2} & 6.5\tiny{$\pm$0.6} & 6.1\tiny{$\pm$0.3} & \textbf{9.1\tiny{$\pm$1.3}} & \textbf{9.1\tiny{$\pm$1.3}} & \textcolor{gray}{9.2\tiny{$\pm$1.4}} \\
 & FF-score $\uparrow$ & 93.5\tiny{$\pm$0.4} & \underline{\textbf{96.0\tiny{$\pm$0.2}}} & 95.7\tiny{$\pm$0.1} & 95.5\tiny{$\pm$0.4} & 95.8\tiny{$\pm$0.2} & 93.8\tiny{$\pm$0.7} & \textbf{93.9\tiny{$\pm$0.7}} & \textcolor{gray}{93.8\tiny{$\pm$0.8}} \\
\midrule
\multirow[c]{4}{*}{\begin{tabular}[c]{@{}l@{}}BIOS-2\\ balanced\\ \end{tabular}} & Fairness $\uparrow$ & 89.7\tiny{$\pm$0.6} & 97.8\tiny{$\pm$0.8} & \underline{\textbf{98.0\tiny{$\pm$0.8}}} & 95.9\tiny{$\pm$0.8} & 96.4\tiny{$\pm$0.3} & \textbf{91.8\tiny{$\pm$2.0}} & 91.6\tiny{$\pm$1.9} & \textcolor{gray}{91.3\tiny{$\pm$1.9}} \\
 & Accuracy $\uparrow$ & 92.4\tiny{$\pm$0.3} & 91.9\tiny{$\pm$0.6} & 91.9\tiny{$\pm$1.5} & 92.6\tiny{$\pm$0.5} & \underline{\textbf{92.9\tiny{$\pm$0.6}}} & 91.9\tiny{$\pm$1.2} & \textbf{92.2\tiny{$\pm$1.0}} & \textcolor{gray}{91.9\tiny{$\pm$1.2}} \\
 & DTO $\downarrow$ & 12.8\tiny{$\pm$0.6} & 8.5\tiny{$\pm$0.4} & 8.4\tiny{$\pm$1.4} & 8.5\tiny{$\pm$0.2} & \underline{\textbf{8.0\tiny{$\pm$0.6}}} & 11.7\tiny{$\pm$1.7} & \textbf{11.5\tiny{$\pm$1.7}} & \textcolor{gray}{12.0\tiny{$\pm$1.6}} \\
 & FF-score $\uparrow$ & 91.1\tiny{$\pm$0.4} & 94.7\tiny{$\pm$0.1} & \underline{\textbf{94.9\tiny{$\pm$0.7}}} & 94.2\tiny{$\pm$0.2} & 94.6\tiny{$\pm$0.3} & 91.8\tiny{$\pm$1.1} & \textbf{91.9\tiny{$\pm$1.2}} & \textcolor{gray}{91.6\tiny{$\pm$1.1}} \\
 \bottomrule
\end{tabular}
}\caption{\label{tab:deb_performance_inlp} Comparison of debiasing methods and selective debiasing using INLP. The best results in the group are in bold, and the best results overall are underlined. The results averaged over 5 random seeds. The gray color corresponds to the results with p-value > 0.05 with respect to standard model.}\end{table*}

\clearpage
\section{Comparison with other Distances}
\label{sec:ablation_dist}
We also conducted additional experiments to compare how proposed selection strategies differ from other similarity measures. Here, we consider several measures, calculated over the output from the last hidden layer of the model, and compare them with SR and KL strategies. The results are presented in \Cref{tab:ablation_dist_leace_last,tab:ablation_dist_leace_cls,tab:ablation_dist_inlp}. In most cases, selection by KL works comparably or better than the best-performing distance-based measure. Moreover, KL scores are easier to compute than distance-based scores. However, in some cases, cosine distance could serve as a replacement for the KL score due to its similar performance.

\begin{table*}[h] 
\footnotesize
\resizebox{\textwidth}{!}{\begin{tabular}{lllllllllllllll}
\toprule
 \textbf{Dataset} & \textbf{Standard} & \textbf{LEACE} & \textbf{SR, 5\%} & \textbf{KL, 5\%} & \textbf{Euclidean, 5\%} & \textbf{Cosine, 5\%} & \textbf{SR, 10\%} & \textbf{KL, 10\%} & \textbf{Euclidean, 10\%} & \textbf{Cosine, 10\%} & \textbf{SR, 15\%} & \textbf{KL, 15\%} & \textbf{Euclidean, 15\%} & \textbf{Cosine, 15\%} \\
  \midrule
MOJI imbalanced & 69.4\tiny{$\pm$0.4} & 71.8\tiny{$\pm$2.6} & 70.8\tiny{$\pm$0.5} & 72.5\tiny{$\pm$0.6} & 71.4\tiny{$\pm$0.7} & 72.2\tiny{$\pm$0.6} & 71.8\tiny{$\pm$0.6} & 73.7\tiny{$\pm$0.7} & 72.5\tiny{$\pm$0.8} & 73.8\tiny{$\pm$0.6} & 72.8\tiny{$\pm$0.5} & 74.1\tiny{$\pm$0.3} & 73.0\tiny{$\pm$0.8} & \textbf{74.3\tiny{$\pm$0.8}} \\
MOJI balanced & 70.7\tiny{$\pm$0.3} & 76.5\tiny{$\pm$2.2} & 72.4\tiny{$\pm$0.1} & 75.4\tiny{$\pm$0.2} & 74.0\tiny{$\pm$0.4} & 75.0\tiny{$\pm$0.2} & 74.0\tiny{$\pm$0.2} & 78.2\tiny{$\pm$0.3} & 76.2\tiny{$\pm$0.4} & 78.1\tiny{$\pm$0.3} & 75.5\tiny{$\pm$0.5} & \textbf{79.8\tiny{$\pm$0.2}} & 77.5\tiny{$\pm$0.4} & 79.2\tiny{$\pm$0.5} \\
BIOS-2 imbalanced & 93.5\tiny{$\pm$0.4} & 72.8\tiny{$\pm$2.3} & \textbf{93.7\tiny{$\pm$1.0}} & 93.3\tiny{$\pm$2.0} & 93.1\tiny{$\pm$1.5} & 92.0\tiny{$\pm$2.0} & \textbf{93.7\tiny{$\pm$1.3}} & 90.3\tiny{$\pm$2.1} & 90.1\tiny{$\pm$1.0} & 88.7\tiny{$\pm$1.5} & 93.1\tiny{$\pm$1.1} & 86.3\tiny{$\pm$2.2} & 86.6\tiny{$\pm$1.5} & 85.6\tiny{$\pm$1.9} \\
BIOS-2 balanced & 91.1\tiny{$\pm$0.4} & 63.0\tiny{$\pm$4.6} & 91.6\tiny{$\pm$1.0} & 92.0\tiny{$\pm$2.1} & 90.3\tiny{$\pm$1.9} & 89.5\tiny{$\pm$1.6} & 92.2\tiny{$\pm$1.9} & 89.6\tiny{$\pm$2.4} & 87.9\tiny{$\pm$3.0} & 86.0\tiny{$\pm$1.7} & \textbf{92.6\tiny{$\pm$2.2}} & 85.4\tiny{$\pm$2.4} & 84.4\tiny{$\pm$1.9} & 82.8\tiny{$\pm$2.1} \\
\bottomrule
\end{tabular}
}\caption{\label{tab:ablation_dist_leace_last} Comparison of FF-score of distance-based scores for LEACE-last for various percentages.}\end{table*}

\begin{table*}[h] \resizebox{\textwidth}{!}{\begin{tabular}{lllllllllllllll}
\toprule
  \textbf{Dataset} & \textbf{Standard} & \textbf{LEACE} & \textbf{SR, 5\%} & \textbf{KL, 5\%} & \textbf{Euclidean, 5\%} & \textbf{Cosine, 5\%} & \textbf{SR, 10\%} & \textbf{KL, 10\%} & \textbf{Euclidean, 10\%} & \textbf{Cosine, 10\%} & \textbf{SR, 15\%} & \textbf{KL, 15\%} & \textbf{Euclidean, 15\%} & \textbf{Cosine, 15\%} \\
  \midrule
MOJI imbalanced & 69.4\tiny{$\pm$0.4} & 70.8\tiny{$\pm$3.0} & 70.8\tiny{$\pm$0.5} & 72.5\tiny{$\pm$0.8} & 71.9\tiny{$\pm$0.5} & 72.2\tiny{$\pm$0.5} & 71.7\tiny{$\pm$0.5} & 74.2\tiny{$\pm$0.6} & 73.3\tiny{$\pm$0.8} & 73.7\tiny{$\pm$1.1} & 72.7\tiny{$\pm$0.4} & \textbf{74.4\tiny{$\pm$0.8}} & 73.9\tiny{$\pm$0.9} & 74.3\tiny{$\pm$1.3} \\
MOJI balanced & 70.7\tiny{$\pm$0.3} & 75.2\tiny{$\pm$2.6} & 72.4\tiny{$\pm$0.1} & 75.5\tiny{$\pm$0.2} & 74.7\tiny{$\pm$0.3} & 74.4\tiny{$\pm$0.5} & 74.0\tiny{$\pm$0.2} & 78.7\tiny{$\pm$0.3} & 77.2\tiny{$\pm$0.4} & 77.3\tiny{$\pm$1.0} & 75.5\tiny{$\pm$0.4} & \textbf{80.0\tiny{$\pm$0.4}} & 78.7\tiny{$\pm$0.4} & 78.9\tiny{$\pm$1.1} \\
BIOS-2 imbalanced & 93.5\tiny{$\pm$0.4} & 69.6\tiny{$\pm$1.7} & 94.6\tiny{$\pm$1.0} & \textbf{95.1\tiny{$\pm$0.4}} & 94.3\tiny{$\pm$0.6} & 94.1\tiny{$\pm$0.6} & 94.7\tiny{$\pm$0.9} & 91.0\tiny{$\pm$1.0} & 89.9\tiny{$\pm$1.2} & 90.3\tiny{$\pm$1.2} & 93.2\tiny{$\pm$0.6} & 85.1\tiny{$\pm$1.1} & 83.9\tiny{$\pm$1.0} & 84.4\tiny{$\pm$1.0} \\
BIOS-2 balanced & 91.1\tiny{$\pm$0.4} & 67.5\tiny{$\pm$3.0} & 92.4\tiny{$\pm$0.6} & 93.9\tiny{$\pm$1.1} & 93.3\tiny{$\pm$1.4} & 92.6\tiny{$\pm$1.1} & 93.7\tiny{$\pm$1.2} & 90.1\tiny{$\pm$2.3} & 87.2\tiny{$\pm$1.8} & 87.0\tiny{$\pm$1.3} & \textbf{94.4\tiny{$\pm$0.9}} & 83.4\tiny{$\pm$2.2} & 80.6\tiny{$\pm$2.0} & 80.9\tiny{$\pm$1.5} \\
\bottomrule
\end{tabular}
}\caption{\label{tab:ablation_dist_leace_cls} Comparison of FF-score of distance-based scores for LEACE-cls for various percentages.}\end{table*}

\begin{table*}[h] \resizebox{\textwidth}{!}{\begin{tabular}{lllllllllllllll}
\toprule
  \textbf{Dataset} & \textbf{Standard} & \textbf{LEACE} & \textbf{SR, 5\%} & \textbf{KL, 5\%} & \textbf{Euclidean, 5\%} & \textbf{Cosine, 5\%} & \textbf{SR, 10\%} & \textbf{KL, 10\%} & \textbf{Euclidean, 10\%} & \textbf{Cosine, 10\%} & \textbf{SR, 15\%} & \textbf{KL, 15\%} & \textbf{Euclidean, 15\%} & \textbf{Cosine, 15\%} \\
  \midrule
  %F-score Euclidean INLP, 5\%
MOJI imbalanced & 69.4\tiny{$\pm$0.4} & \textbf{71.9\tiny{$\pm$2.2}} & 69.5\tiny{$\pm$0.3} & 71.0\tiny{$\pm$0.8} & 69.5\tiny{$\pm$0.5} & 69.5\tiny{$\pm$0.5} & 69.6\tiny{$\pm$0.5} & 71.8\tiny{$\pm$1.0} & 69.7\tiny{$\pm$0.7} & 69.6\tiny{$\pm$0.6} & 70.1\tiny{$\pm$0.5} & \textbf{71.9\tiny{$\pm$1.3}} & 69.8\tiny{$\pm$0.9} & 69.7\tiny{$\pm$0.8} \\
MOJI balanced & 70.7\tiny{$\pm$0.3} & 71.9\tiny{$\pm$3.2} & 71.2\tiny{$\pm$0.1} & 72.0\tiny{$\pm$0.6} & 70.9\tiny{$\pm$0.4} & 71.1\tiny{$\pm$0.5} & 71.6\tiny{$\pm$0.4} & 72.5\tiny{$\pm$0.9} & 71.2\tiny{$\pm$0.5} & 71.3\tiny{$\pm$0.6} & 71.8\tiny{$\pm$0.5} & \textbf{72.8\tiny{$\pm$1.0}} & 71.4\tiny{$\pm$0.6} & 71.5\tiny{$\pm$0.6} \\
BIOS-2 imbalanced & 93.5\tiny{$\pm$0.4} & 93.8\tiny{$\pm$0.7} & \textbf{93.9\tiny{$\pm$0.7}} & 93.8\tiny{$\pm$0.8} & 93.5\tiny{$\pm$0.4} & 93.5\tiny{$\pm$0.4} & 93.8\tiny{$\pm$0.7} & 93.8\tiny{$\pm$0.8} & 93.5\tiny{$\pm$0.4} & 93.5\tiny{$\pm$0.4} & 93.8\tiny{$\pm$0.7} & 93.8\tiny{$\pm$0.8} & 93.5\tiny{$\pm$0.4} & 93.5\tiny{$\pm$0.4} \\
BIOS-2 balanced & 91.1\tiny{$\pm$0.4} & 91.8\tiny{$\pm$1.1} & 91.5\tiny{$\pm$0.9} & 91.6\tiny{$\pm$1.1} & 91.1\tiny{$\pm$0.4} & 91.1\tiny{$\pm$0.4} & 91.8\tiny{$\pm$1.1} & 91.7\tiny{$\pm$1.1} & 91.1\tiny{$\pm$0.4} & 91.1\tiny{$\pm$0.4} & \textbf{91.9\tiny{$\pm$1.2}} & 91.7\tiny{$\pm$1.2} & 91.1\tiny{$\pm$0.4} & 91.1\tiny{$\pm$0.4} \\
\bottomrule
\end{tabular}
}\caption{\label{tab:ablation_dist_inlp} Comparison of FF-score of distance-based scores for INLP for various percentages.}\end{table*}

\clearpage

\section{Computational Efficiency}
\label{sec:comp_stats}
To estimate the computational efficiency of selective debiasing, we calculated the inference time of the standard model and the model with selective debiasing. The results are presented in \Cref{tab:inference_time,tab:inference_time_ratios}. \Cref{tab:inference_time} shows the inference time of models averaged for 10 runs, while \Cref{tab:inference_time_ratios} presents computational overhead for each debiasing method. The computational overhead is calculated as follows:
\begin{equation}
    CompOverhead=100 \cdot \left( \frac{T_{selective}}{T_{standard}} - 1\right),
\end{equation}
where $T$ is the summary inference time of the debiasing method for all datasets. These experiments were conducted on one Nvidia H100 GPU. The proposed selective debiasing approach does not introduce much computational overhead -- for LEACE-last and LEACE-cls it is less than 1\%.

Table \ref{tab:deb_methods} shows a detailed comparison of debiasing methods. As one can see, at-training and pre-processing debiasing methods require a training model from scratch, while post-processing methods with selective debiasing do not require this. Hence, post-processing methods are especially beneficial when the full dataset or the model is unavailable, while selective debiasing allows for increasing the overall performance of these methods. On the other hand, there is some computational overhead for post-processing methods compared to other ones. However, this overhead is negligible in most cases.

\begin{table*}[h] 
\footnotesize
\centering\resizebox{0.8\textwidth}{!}{\begin{tabular}{l|ll|ll|ll}

\toprule
 \multirow{2}{*}{\textbf{Dataset}} & \multicolumn{2}{c|}{\textbf{LEACE-last}} & \multicolumn{2}{c|}{\textbf{LEACE-cls}} & \multicolumn{2}{c}{\textbf{INLP}} \\
  & \textbf{Selective} & \textbf{Standard} & \textbf{Selective} & \textbf{Standard} & \textbf{Selective} & \textbf{Standard} \\
 \midrule
MOJI imbalanced & 3.738\tiny{$\pm$0.011} &  3.737\tiny{$\pm$0.020}   & 3.762\tiny{$\pm$0.009}     &   3.749\tiny{$\pm$0.035}    & 3.775\tiny{$\pm$0.008}    &     3.730\tiny{$\pm$0.008}\\
MOJI balanced & 5.978\tiny{$\pm$0.023}    &     5.96\tiny{$\pm$0.014}    & 6.008\tiny{$\pm$0.014}    &    5.971\tiny{$\pm$0.024}   & 6.053\tiny{$\pm$0.017}    &    5.974\tiny{$\pm$0.016}\\
BIOS-2 imbalanced & 6.064\tiny{$\pm$0.018}     &   6.059\tiny{$\pm$0.033}    & 6.090\tiny{$\pm$0.018}    &    6.049\tiny{$\pm$0.013}    & 6.116\tiny{$\pm$0.022}   &     6.051\tiny{$\pm$0.022}\\
BIOS-2 balanced & 1.526\tiny{$\pm$0.007}    &    1.525\tiny{$\pm$0.006}    & 1.544\tiny{$\pm$0.024}     &   1.527\tiny{$\pm$0.025}    & 1.542\tiny{$\pm$0.004}     &   1.525\tiny{$\pm$0.004}\\

\bottomrule
\end{tabular}
}\caption{\label{tab:inference_time} Inference time of standard model and model with applied selective debiasing (in seconds, averaged for 10 runs).}\end{table*}
\begin{table}[h] 
\footnotesize
\centering\resizebox{0.4\textwidth}{!}{\begin{tabular}{l|l|l|l}

\toprule
 & \textbf{LEACE-last} & \textbf{LEACE-cls} & \textbf{INLP} \\
 \midrule
Overhead, \% & 0.14 & 0.62 & 1.19\\
\bottomrule
\end{tabular}
}\caption{\label{tab:inference_time_ratios} The computational overhead of selective debiasing for various methods.}\end{table}
\begin{table*}[h] 
\footnotesize
\centering\resizebox{0.8\textwidth}{!}{\begin{tabular}{l||l|ll|ll||lll}
\toprule
\textbf{Debiasing method type} & \textbf{Base} & \multicolumn{2}{c|}{\textbf{At-training}} & \multicolumn{2}{c||}{\textbf{Pre-processing}} & \multicolumn{3}{c}{\textbf{Selective}} \\
\toprule
 \textbf{Debiasing method} & \textbf{Standard} & \textbf{Adv} & \textbf{DAdv} & \textbf{BTEO} & \textbf{BTJ} & \begin{tabular}[l]{@{}l@{}}\textbf{LEACE-last}\\ \textbf{selective}\end{tabular} & \begin{tabular}[l]{@{}l@{}}\textbf{LEACE-cls}\\ \textbf{selective}\end{tabular} & \begin{tabular}[l]{@{}l@{}}\textbf{INLP}\\ \textbf{selective}\end{tabular} \\
 \midrule
 \begin{tabular}[l]{@{}l@{}}Require model retraining\\from Standard model\end{tabular} & \ding{53} & \ding{51} & \ding{51} & \ding{51} & \ding{51} & \ding{53} & \ding{53} & \ding{53}\\

  At-training method & \ding{53} & \ding{51} & \ding{51} & \ding{53} & \ding{53} & \ding{53} & \ding{53} & \ding{53}\\
  Pre-processing method & \ding{53} & \ding{53} & \ding{53} & \ding{51} & \ding{51} & \ding{53} & \ding{53} & \ding{53}\\
  Post-processing method & \ding{53} & \ding{53} & \ding{53} & \ding{53} & \ding{53} & \ding{51} & \ding{51} & \ding{51}\\
  \begin{tabular}[l]{@{}l@{}}Inference speed\\ (relative to Standard model)\end{tabular} & 1.000 & 1.000 & 1.000 & 1.000 & 1.000 & 1.001 & 1.006 & 1.012 \\
\bottomrule
\end{tabular}
}\caption{\label{tab:deb_methods} Debiasing methods comparison. At-training and pre-processing debiasing methods can have the same inference speed, but require model training from scratch, which is impossible in some cases.}\end{table*}

\clearpage

\end{document}